\documentclass[letterpaper]{article} 
\usepackage{aaai2027}  
\usepackage[hyphens]{url}  
\usepackage{graphicx} 
\usepackage{natbib}  
\usepackage{caption} 
\usepackage{algorithm}
\usepackage{algorithmic}
\usepackage{amssymb}
\usepackage{amsmath}
\usepackage{bm}
\usepackage{newfloat}
\usepackage{listings}
\DeclareCaptionStyle{ruled}{labelfont=normalfont,labelsep=colon,strut=off} 
\floatstyle{ruled}
\newfloat{listing}{tb}{lst}{}
\floatname{listing}{Listing}
\usepackage{makecell}

\newcommand{\method}[2]{\makecell[l]{#1\\[-1pt]\citep{#2}}}
\usepackage{booktabs}

\newcommand{\ourmethod}{VLA-Talker}
\newcommand{\spatialtag}[1]{\texttt{<spatial>}#1\texttt{</spatial>}}

\title{In-Context VLA: Endowing Vision-Language-Action Models with Language via In-Context Post-Training and Agentic Tool Use}
\author {
Jiarui Yang\textsuperscript{\rm 1,\rm 2}, Wen Huang\textsuperscript{\rm 2}, Jiale Zhang \textsuperscript{\rm 2}, Maowei Hu \textsuperscript{\rm 2}, Hang Guo\textsuperscript{\rm 3}
}
\affiliations {
    \textsuperscript{\rm 1} Nankai University 
    \textsuperscript{\rm 2} Tsinghua University \\
    \textsuperscript{\rm 3} Swiss Federal Institute of Technology in Lausanne (EPFL) 
}

\begin{document}

\maketitle

\begin{abstract}
Vision-Language-Action (VLA) models have become the dominant recipe for generalist manipulation, yet they are almost universally trained by behavior cloning: a policy imitates expert action chunks conditioned on a static image and a fixed instruction. A natural remedy is to inject explicit reasoning through textual chain-of-thought (CoT). We show, both empirically and analytically, that free-form textual CoT degrades low-level control: the reasoning it produces is ungrounded, its latency breaks closed-loop timing, and, crucially, the reasoning and action tokens are optimized against conflicting objectives so that the policy learns to narrate rather than to act. We argue that what a VLA needs is not the ability to generate language, but the ability to consume grounded language. To this end we introduce \textbf{\ourmethod{}}, a framework that endows a VLA with language competence through (i) in-context post-training, in which perceptual evidence is injected as structured context and the model is supervised only on actions, and (ii) an agentic tool-use interface, in which the policy queries open-vocabulary detectors, monocular depth, and a vision--language model to actively acquire task-relevant information. Rather than emitting a single templated caption, our data engine produces diverse, paraphrased, and evidence-conditioned spatial descriptions, so that the policy learns to interpret language it has never seen verbatim. Across the RoboCasa-GR1, SimplerEnv, and LIBERO simulation benchmarks, together with 8 real-world robot manipulation tasks, our method consistently achieves SOTA results in both performance and efficiency when compared with CoT-based approaches under matched configurations.
\end{abstract}



\begin{figure}[t]
    \centering

    \includegraphics[width=0.98\columnwidth]{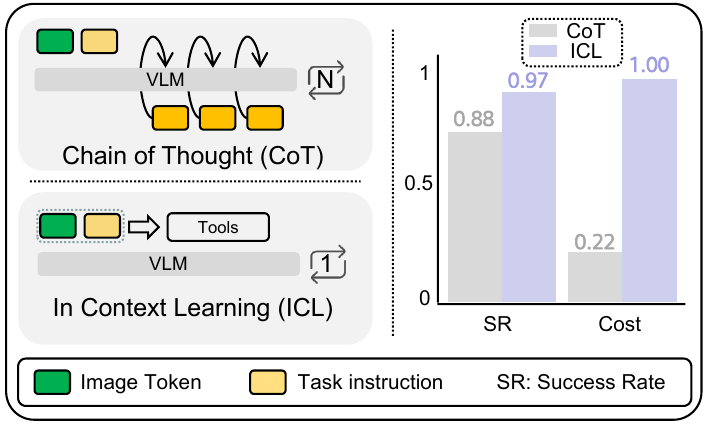}
    \caption{Comparison between free-form CoT generation and ICL for VLA control. CoT requires the VLM to autoregressively generate N reasoning tokens before acting, while ICL/tool-use injects structured evidence in a single pass. The bar chart shows that ICL matches or exceeds CoT's success rate (SR) while incurring far lower inference cost}
    \label{fig:teaser}
\end{figure}

\section{Introduction}


Vision-Language-Action (VLA) models have rapidly become the default recipe for building generalist robot policies \citep{ma2026survey}. By fine-tuning a pretrained vision-language backbone to emit low-level action tokens, systems such as OpenVLA \citep{kim2024openvla} and its optimized-fine-tuning variants achieve strong performance across simulated and real manipulation suites. Almost without exception, however, these models are trained by behavior cloning: given a static observation $o_t$ and a fixed natural-language instruction $\ell$, the policy $\pi_\theta(a_{t:t+H} \mid o_t, \ell)$ is optimized to reproduce expert action chunks. Behavior cloning is attractive because it is simple and data-efficient, but it also inherits two well-known weaknesses. First, the instruction $\ell$ is treated as an opaque conditioning string that is memorized rather than understood: paraphrasing the instruction, or referring to an object by an unseen synonym, routinely degrades success \citep{liu2023libero}. Second, the observation is consumed in a single feed-forward pass, so the policy cannot actively seek information it is missing---it must guess where the target is even when the answer is one look away \citep{ahn2022can}.


A natural hypothesis is that adding explicit reasoning should help. The community has therefore explored chain-of-thought (CoT) for VLAs, in which the model first generates a textual (or ``thinking-with-image'') rationale and then predicts actions \citep{zhao2025cot, wang2026vla}. We find that free-form generative CoT is at best neutral and often harmful for low-level control. We attribute this to three coupled failures. (1) Grounding gap: the rationale is generated from the same static features that the action head already sees, so it adds no new evidence; when it hallucinates a location, it actively misleads the action head \citep{lin2026systematic}. (2) Objective interference: token-level supervision on the rationale competes with action supervision inside a single autoregressive loss, and because language tokens vastly outnumber the few action tokens, the policy is pushed to become a fluent narrator rather than an accurate actor. (3) Latency and drift: generating hundreds of rationale tokens per decision breaks the closed-loop timing that manipulation requires, and any error in the long autoregressive prefix propagates into the action suffix. 


Our central claim is a shift in perspective: a VLA does not need the ability to generate language; it needs the ability to consume grounded language. Human operators rarely verbalize a paragraph before every motion; instead, they look, acquire the specific fact they lack (``the mug is to the left of the plate, slightly below the gripper''), and act \citep{kojima2022large}. We therefore decouple the two roles that generative CoT conflates: acquiring evidence and using evidence. Evidence acquisition is delegated to an external agentic tool-use loop---open-vocabulary detection, monocular depth, and a vision-language model (VLM \citep{liu2023visual})---while the VLA policy is trained only to read the resulting evidence and to act on it.



As shown in Fig.~\ref{fig:teaser}, we instantiate this idea in \textbf{\ourmethod{}}. At its core is
in-context post-training \citep{sirko2025dip}: on keyframes, structured perceptual evidence
is injected into the prompt inside \spatialtag{$\cdots$} tags, and the policy is supervised only on the action tokens. The model thus never learns to write the evidence---it only learns to condition on it, which sidesteps objective interference and latency entirely. In contrast, CoT-based methods require iterative autoregressive generation of textual reasoning during inference, whereas our approach queries external, more powerful APIs for information acquisition, achieving nearly a 4.6$\times$ speedup in inference efficiency. Evidence is produced by an agentic pipeline that projects the gripper into image space analytically, localizes target objects with a detector, samples relative depth, and falls back to a VLM when detection is uncertain. Crucially, to prevent the policy from overfitting to a single phrasing, our data engine renders diverse spatial descriptions: coordinates, relative offsets, depth comparisons, and directional hints are paraphrased and recombined so that the same geometric fact appears under many surface forms. This is exactly the ``language competence'' a VLA lacks under plain behavior cloning---robustness to how information is expressed. Finally, we further fine-tune the in-context policy with a trajectory-level reinforcement-learning objective (GRPO with a sparse success reward \citep{shao2024deepseekmath}). This stage aligns when the policy invokes tools with true task success, closing the gap between per-token imitation and trajectory-level outcomes.

Across three simulation benchmarks, RoboCasa-GR1 \citep{bjorck2025gr00t}, SimplerEnv \citep{simpler}, and LIBERO \citep{liu2023libero}, our method achieves SOTA performance compared with existing SOTA approaches and CoT-based baselines. Compared with standard behavior cloning baselines, the proposed in-context reasoning and tool-use framework improves the task success rate by nearly 10\%. Furthermore, extensive real-world evaluations on an Agibot \citep{bu2025agibot} humanoid platform across eight tabletop manipulation tasks demonstrate the superiority and effectiveness of our approach.

\section{Related Work}
\paragraph{Chain-of-thought reasoning for VLAs.}
Chain-of-thought (CoT) prompting has been adapted to vision-language-action models in several forms. CoT-VLA~\citep{zhao2025cot} inserts an intermediate visual rationale by autoregressively predicting future subgoal images before emitting actions, improving temporal planning on both simulated and real manipulators. Subsequent work explores textual rationales, dual visual-linguistic streams~\citep{zhong2026dualcot}, action-space deliberation (ACoT)~\citep{zhong2026acot}, and trajectory-level planning with failure recovery~\citep{li2026tcot}. VLA-Thinker~\citep{wang2026vla} further treats perception itself as a dynamically invocable reasoning step. While these approaches demonstrate gains on high-level or long-horizon tasks, free-form generative CoT often incurs substantial latency, compounds autoregressive error, and can interfere with low-level control objectives; several recent studies therefore move reasoning into latent space to mitigate these costs~\citep{bai2026latent}. 

\paragraph{Tool use and agentic VLAs.}

Recent work explores tool-augmented and agentic extensions of VLAs to improve perception, reasoning, and long-horizon manipulation. LLaVA-Plus~\citep{liu2024llava} maintains a skill repository of vision tools and trains multimodal agents to activate and compose them on the fly. MemoryVLA~\citep{shi2025memoryvla} and RoboMemory~\citep{lei2025robomemory} equip agents with external perceptual-cognitive and multi-type memory systems for temporal reasoning. Long-VLA~\citep{fanlong} further enhances skill chaining via phase-aware masking, while VisTA~\citep{huang2025visualtoolagent} uses reinforcement learning for adaptive tool selection. SpaceTools~\citep{chen2025spacetools} trains VLM-based agents to invoke external
perception and robot-control tools for spatial reasoning and manipulation.
VLAs-as-Tools~\citep{lei2026towards} decomposes long-horizon tasks by organizing
specialized VLAs as callable subroutines coordinated by a high-level agent. Beyond these approaches, recent embodied agent frameworks explore reusable skill libraries and modular tool interfaces to improve robustness and generalization in interactive environments.

\begin{figure*}[t]
    \centering
    \includegraphics[width=0.98\textwidth]{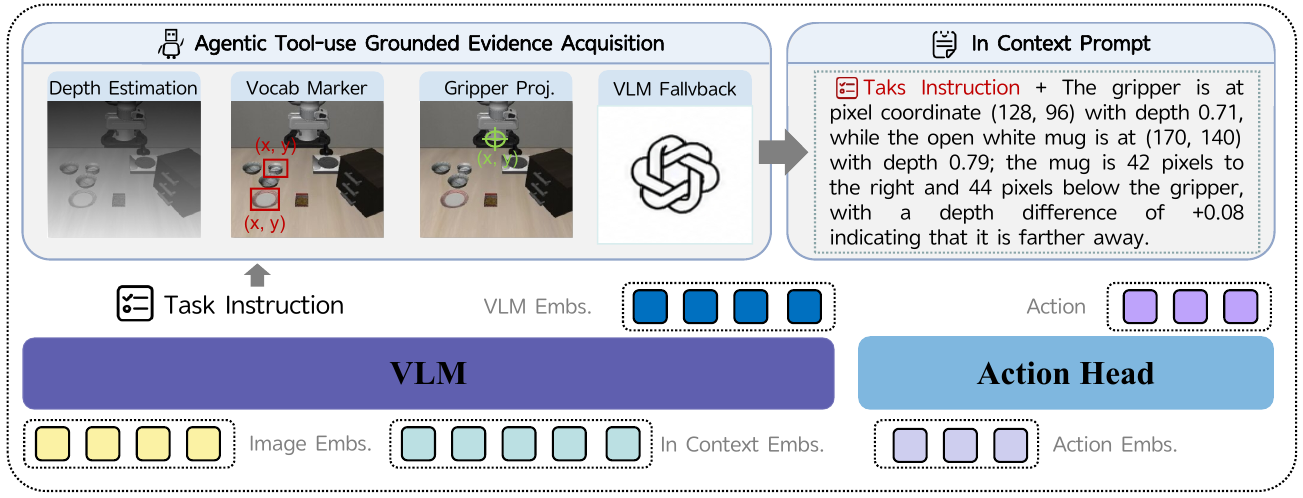}
    \caption{Overview of the VLA-Talker framework. An agentic tool-use module (depth estimation, open-vocabulary object detection, gripper projection, and VLM fallback) acquires grounded spatial evidence at each keyframe, which is rendered into a diverse, paraphrased in-context prompt. This evidence is injected into the VLM alongside the task instruction and image.}
    \label{fig:method}
\end{figure*}

\section{Method}

\subsection{Problem Setup and Notation}

We consider language-conditioned manipulation. At timestep $t$ the policy observes an RGB image $o_t \in \mathbb{R}^{H\times W\times 3}$ (optionally a wrist view), proprioception $s_t$, and a fixed instruction $\ell$. It must emit an action chunk $a_{t:t+H}$, where each action is a 7-DoF end-effector command (a translation, a rotation, and a gripper signal). A standard VLA is trained by behavior cloning,
\begin{equation}
\mathcal{L}_{\mathrm{BC}}(\theta)
= -\!\!\sum_{k=0}^{H-1}\!\log \pi_\theta\!\big(a_{t+k}\mid o_t, s_t, \ell,
a_{t:t+k}\big),
\label{eq:bc}
\end{equation}
which imitates expert chunks but never teaches the policy to seek the information it lacks or to interpret unfamiliar phrasings of $\ell$.

\subsection{Why Generative CoT Hurts Low-Level Control}

Generative CoT augments Eq.~\eqref{eq:bc} with a rationale $r$ produced by the same policy, giving a combined objective
\begin{equation}
\mathcal{L}_{\mathrm{CoT}}
= \underbrace{-\!\sum_j \log \pi_\theta(r_j \mid \cdot)}_{\text{language}}
\;+\; \underbrace{-\!\sum_k \log \pi_\theta(a_{t+k}\mid r, \cdot)}_{\text{action}} .
\label{eq:cot}
\end{equation}
Three problems follow. (1) The rationale $r$ is generated from the same
features $o_t$ available to the action head, so in expectation it carries no new information; when it is wrong, it becomes a noisy, misleading prefix \citep{lin2026systematic}. (2) The language term contains orders of magnitude more tokens than the action term, so gradient mass is dominated by ``sounding reasonable,'' and the policy drifts toward narration. (3) Sampling $r$ autoregressively at inference adds hundreds of tokens of latency per decision and lets any early mistake contaminate the action suffix. The takeaway motivates our design: \emph{remove the language term from the loss, and replace self-generated $r$ with externally grounded evidence.}

\subsection{Overview of \ourmethod{}}

\ourmethod{} keeps the VLA backbone and action head unchanged and adds two components (Fig.~\ref{fig:method}): (i) an agentic tool-use loop that acquires grounded spatial evidence, and (ii) an in-context post-training scheme that injects this evidence as read-only context and supervises only actions. The combined training objective is simply Eq.~\eqref{eq:bc} with the observation extended by an evidence context $c_t$,
\begin{equation}
\mathcal{L}(\theta)
= -\!\!\sum_{k=0}^{H-1}\!\log \pi_\theta\!\big(a_{t+k}\mid o_t, s_t, \ell, c_t,
a_{t:t+k}\big),
\label{eq:ours}
\end{equation}
where $c_t$ is produced by the tool loop on keyframes and is empty otherwise. No token of $c_t$ ever appears in the loss.

\subsection{Agentic Tool Use for Grounded Evidence}

At each keyframe the policy delegates evidence acquisition to a small set of perception tools, orchestrated as an agent that answers a single query: \emph{what are the image-space locations and relative depths of the gripper and the task-relevant objects?}

\paragraph{Depth and relative geometry.}
A monocular depth estimator produces a depth map normalized to $[0,1]$ ($1{=}$near, $0{=}$far). We sample relative depth at the gripper and object pixels, which lets the agent reason about height ordering (\emph{is the object above or below the current gripper height?}) without metric calibration.

\paragraph{Object localization (open-vocabulary + fallback).}
Target and destination object names are parsed from the instruction (e.g., ``put the white mug on the left plate'' $\rightarrow$ $\{$white mug, left plate$\}$). An open-vocabulary detector localizes each object, returning a pixel centroid $(u_o, v_o)$. When the detector is uncertain or fails, an agentic fallback queries a VLM locator that returns either approximate coordinates or a qualitative spatial description. This detector-then-VLM cascade mirrors how tool-using agents escalate to a more general (but noisier) tool only when the specialized one is insufficient.

\paragraph{Gripper localization.}
Because the end-effector world position is known from proprioception ($s_t[0{:}3]$), we project it into image space using the per-task camera intrinsics and extrinsics extracted from the simulator, correcting for the renderer's vertical flip. This yields an exact gripper pixel $(u_g, v_g)$ up to sub-pixel projection error---no learning required.

\paragraph{Evidence tuple.}
The tool loop thus emits, per keyframe, a structured evidence tuple $c_t^{\mathrm{raw}} = \big(\,(u_g,v_g,d_g,\text{grip state}),\; \{(u_o,v_o,d_o)\}_{o},\; \text{relations}\,\big)$, where relations capture gripper$\leftrightarrow$object and object$\leftrightarrow$object offsets and depth comparisons. This tuple is the ground truth of what the policy is allowed to know.

\subsection{Caption Rendering}

A naive renderer converts the raw evidence $c_t^{\mathrm{raw}}$ into a single fixed template, e.g. always emitting ``\texttt{gripper: (x=U, y=V, depth=D); mug: (x=..)}''. Training on one template teaches the policy to pattern-match a surface string rather than to understand spatial language, so any deviation in phrasing at test time collapses performance. To avoid this, \ourmethod{} instead renders the same evidence tuple into diverse, paraphrased realizations---varying reference modality, lexical form, and depth verbalization---all wrapped in \texttt{<spatial>$\cdots$</spatial>} tags. Holding the geometric meaning fixed while varying the surface form forces the policy to learn a mapping from diverse language to the same grounded intent. We defer the full description of the rendering axes and worked examples to Appendix.

\begin{listing}[t]
\begin{lstlisting}[
numbers=none,
basicstyle=\ttfamily\footnotesize,
aboveskip=0pt,
belowskip=0pt
]
# Realization A (coordinate-rich, detector succeeded)
<spatial>
The gripper is at (128,96), depth 0.71; the open white mug
is at (170,140), depth 0.79. The mug is 42 px right and
44 px below the gripper, with +0.08 depth difference
indicating that it is farther away.
</spatial>

# Realization B (paraphrased, VLM fallback)
<spatial>
My open gripper sits upper-center. The white mug lies
lower-right, slightly farther on the table, and below the
gripper's current height.
</spatial>
\end{lstlisting}
\caption{Two diverse renderings of the same evidence tuple.
The policy is supervised only on the actions that follow, never on this text.}
\label{lst:spatial}
\end{listing}

\subsection{In-Context Post-Training}

Given the rendered context $c_t$, we form the training sequence
$[\,\ell,\; \langle\text{image}\rangle,\; c_t,\; a_{t:t+H}\,]$ and optimize
Eq.~\eqref{eq:ours}. The defining property is the supervision mask: the
loss is applied only to the action tokens (and the single separator that
precedes them), with all instruction and \texttt{<spatial>} tokens masked out.
Therefore the model never learns to produce $c_t$; it only learns to
attend to it when predicting actions. This eliminates the objective
interference of Eq.~\eqref{eq:cot}, incurs no autoregressive generation latency and cannot
drift into narration.

\paragraph{Keyframe gating.}
Injecting evidence at every step is wasteful and unnecessary. We inject $c_t$
only on keyframes---the initial frame, frames where the gripper
open/close state changes, and periodic progress checks---and leave $c_t$ empty
otherwise. On empty-context steps the model behaves as a fast vanilla VLA. At
inference the same schedule is applied: the tool loop runs only on keyframes,
so the amortized overhead is small.

\subsection{Trajectory-Level RL for Causal Alignment}
\label{sec:rl}

As the final stage of our method, we align complete
perception--reasoning--action trajectories to sparse task-level rewards using
Group Relative Policy Optimization (GRPO \citep{shao2024deepseekmath}). Because the interface is unchanged
(evidence is still injected, actions are still the only produced tokens), this
stage composes cleanly on top of the in-context checkpoint.

\paragraph{Trajectory and reward.}
We treat \ourmethod{} as a stochastic policy $\pi_\theta$ over multimodal
trajectories. Given an instruction $\ell$ and initial observation $V_0$, a
rollout produces
\begin{equation}
\tau = \{\,T_1, C_1, V_1, \dots, T_n, A_n\,\},
\label{eq:traj}
\end{equation}
where, in Eq.~\eqref{eq:traj}, $T_k$ is a reasoning/decision step, $C_k$ a perception tool call, $V_k$
the returned evidence (the injected \spatialtag{} context), and $A_n$ the
executed action chunk. The reward is sparse, assigned only at the end of
the episode from a verifiable success signal, plus a small format-regularization
term that keeps the tool-call syntax well-formed:
\begin{equation}
R(\tau) = \alpha_s\, \mathbb{I}_{\text{success}} + \alpha_f\, \mathbb{I}_{\text{format}},
\label{eq:reward}
\end{equation}
with $\mathbb{I}_{\text{success}}=1$ iff the task is completed and
$\mathbb{I}_{\text{format}}=1$ iff the emitted tool calls parse against the
schema. No dense or intermediate rewards are used, so the policy is free to
discover when to call tools rather than being told.

\paragraph{Group-relative advantage.}
For each $(\ell, V_0)$ we sample a group of $M$ trajectories
$\{\tau_1,\dots,\tau_M\}\sim\pi_\theta$ and estimate each trajectory's advantage
by normalizing its reward within the group, which removes the need for a
learned value function and sharply reduces variance under sparse feedback:
\begin{equation}
A_i = \frac{R(\tau_i) - \mathrm{mean}\big(\{R(\tau_j)\}_{j=1}^{M}\big)}
            {\mathrm{std}\big(\{R(\tau_j)\}_{j=1}^{M}\big)}.
\label{eq:adv}
\end{equation}

\paragraph{Objective.}
Following the group-relative formulation, we optimize a clipped surrogate with a
KL anchor to the reference (in-context) policy $\pi_{\text{ref}}$, as given in
Eq.~\eqref{eq:grpo}:
\begin{equation}
\begin{aligned}
J(\theta) = \mathbb{E}\Big[ \tfrac{1}{M}\!\sum_{i=1}^{M} \big(
& \min( r_i A_i,\; \mathrm{clip}(r_i, 1{-}\epsilon, 1{+}\epsilon)\,A_i ) \\
& - \beta\, D_{\mathrm{KL}}(\pi_\theta \,\|\, \pi_{\text{ref}}) \big) \Big],
\end{aligned}
\label{eq:grpo}
\end{equation}
where $r_i(\theta)=\pi_\theta(\tau_i\mid\ell)/\pi_{\theta_{\text{old}}}(\tau_i\mid\ell)$
is the importance ratio, $\epsilon$ the clipping range, and $\beta$ the KL
weight. Only the action tokens contribute to the ratio; injected evidence tokens
remain non-generated, so RL refines how and when to act on evidence
without ever turning the model back into a rationale generator.
\begin{table}[t]
\centering
\caption{LIBERO simulation results on four task suites. We report success
rates (\%) on Spatial, Object, Goal, and Long, together with the average across
the four suites. Best in \textbf{bold}.}
\label{tab:main}
\setlength{\tabcolsep}{4pt}

\scalebox{0.79}{
\begin{tabular}{lccccc}
\toprule
Method & L-Spatial & L-Object & L-Goal & L-Long & Avg. \\
\midrule
\method{\textbf{Diffusion Policy}}{chi2025diffusion} 
& 78.5 & 87.5 & 73.5 & 64.8 & 76.1 \\

\method{\textbf{OpenVLA}}{kim2024openvla}         
& 84.7 & 88.4 & 79.2 & 53.7 & 76.5 \\

\method{\textbf{SpatialVLA}}{qu2025spatialvla}       
& 88.2 & 89.9 & 78.6 & 55.5 & 78.1 \\

\method{\textbf{CoT-VLA}}{zhao2025cot}        
& 87.5 & 91.6 & 87.6 & 69.0 & 83.9 \\

\method{\textbf{GR00T N1}}{bjorck2025gr00t}         
& 94.4 & 97.6 & 93.0 & 90.6 & 93.9 \\

\method{\textbf{F1}}{lv2025f1}             
& 98.2 & 97.8 & 95.4 & 91.3 & 95.7 \\

\method{\textbf{InternVLA-M1}}{chen2025internvla}     
& 98.0 & 99.0 & 93.8 & 92.6 & 95.9 \\

\method{\textbf{$\bm{\pi}_0$}}{black2024pi_0}       
& 98.0 & 96.8 & 94.4 & 88.4 & 94.4 \\

\method{\textbf{$\bm{\pi}_{0.5}$}}{intelligence2025pi_}     
& \textbf{98.8} & 98.2 & 98.0 & 92.4 & 96.9 \\

\method{\textbf{VLA-Thinker}}{wang2026vla}    
& 97.7 & 98.5 & 97.5 & \textbf{94.4} & 97.0 \\

\midrule

\textbf{Gen-CoT}
& 97.5 & 98.6 & 97.2 & 91.6 & 96.2 \\

\textbf{\ourmethod{}}
& 98.2 & \textbf{99.2} & \textbf{98.4} &
93.6 & \textbf{97.4} \\

\bottomrule
\end{tabular}
}

\end{table}

\begin{table}[t]
\centering
\caption{RoboCasa-GR1 simulation results on four representative pick-and-place
tasks. Best in \textbf{bold}.}
\label{tab:robocasa}
\setlength{\tabcolsep}{4pt}
\scalebox{0.8}{
\begin{tabular}{lccccc}
\toprule
Method & PnP Bottle & PnP Can & PnP Cup & PnP Milk & Avg. \\
\midrule

\method{\textbf{GR00T N1.5}}{bjorck2025gr00t}
& 54.0 & 50.0 & 38.0 & \textbf{60.0} & 48.2 \\

\method{\textbf{TwinBrainVLA}}{yu2026twinbrainvla}
& 74.0 & 72.0 & 52.0 & 60.0 & 54.6 \\

\method{\textbf{PhysBrain}}{lin2025physbrain}
& 74.0 & 68.0 & 42.0 & 54.0 & 50.0 \\

\method{\textbf{LangForce}}{lian2026langforce}
& 72.0 & 78.0 & 46.0 & 56.0 & 52.6 \\

\method{\textbf{ABot-M0}}{yang2026abot}
& \textbf{86.0} & 74.0 & 48.0 & 46.0 & 58.3 \\

\midrule

\textbf{Gen-CoT} 
& 48.0 & 76.0 & \textbf{52.0} & 50.0 & 46.5 \\

\textbf{\ourmethod{}} 
& 76.0 & \textbf{78.0} & 48.0 & 58.0 & \textbf{59.5} \\

\bottomrule
\end{tabular}
}
\end{table}

\section{Experiments}

\paragraph{Benchmarks.}
We evaluate on three settings that span diverse embodiments and scenes.
\textbf{(i) RoboCasa-GR1} \citep{bjorck2025gr00t}: a large-scale kitchen manipulation benchmark on the
GR1 humanoid with diverse scenes, objects, and long-horizon activities; we report
per-task success rate (SR, \%) on representative pick-and-place tasks and the
average over all $24$ tasks. \textbf{(ii) SimplerEnv} \citep{simpler}: a WidowX manipulation
suite; we report SR on four held-out tasks and their average, stressing
generalization to unseen tabletop configurations. \textbf{(iii) LIBERO} \citep{liu2023libero}: four
language-conditioned suites (LIBERO-Spatial, -Object, -Goal, -Long), each
evaluated under $50$ randomized initial conditions per task and reported as
SR averaged within and across suites.

\paragraph{Backbone and baselines.}
We initialize from publicly available OpenVLA-OFT \citep{kim2024openvla} weights and train with parallel decoding and action chunking. During training and inference we use only a single third-person RGB view and the language instruction (no wrist camera). We compare against a broad set of published VLA systems on each benchmark, and report a generative-CoT (matched
evidence) variant, denoted \textsc{Gen-CoT}, that uses the identical evidence
tuples from the same tool loop but generates and supervises them as text
rather than injecting them. For fairness, this variant and
\ourmethod{} differ only in whether evidence is generated or injected, and in
what is supervised.

\paragraph{Implementation.}
Tools are an open-vocabulary detector (GroundingDino \citep{liu2024grounding}), a depth estimator (DepthAnything \citep{yang2024depth}), and a VLM
locator fallback (Qwen2.5-VL-7B). Keyframes follow the
initial/gripper-change/periodic schedule of the method section. Training uses the
two-stage recipe of Section~\ref{sec:rl}: a supervised in-context cold-start
followed by GRPO. We use a batch size of $64$ for the in-context stage and $128$
for RL, learning rates $1{\times}10^{-5}$ and $2{\times}10^{-6}$ respectively,
optimized with AdamW; RL samples $M$ trajectories per prompt with group-relative
advantage (Eq.~\eqref{eq:adv}), sparse success reward with a small format term
(Eq.~\eqref{eq:reward}), and a KL anchor to the in-context checkpoint. Details are in the Appendix.

\begin{figure}[t]
    \centering
    \includegraphics[width=0.98\columnwidth]{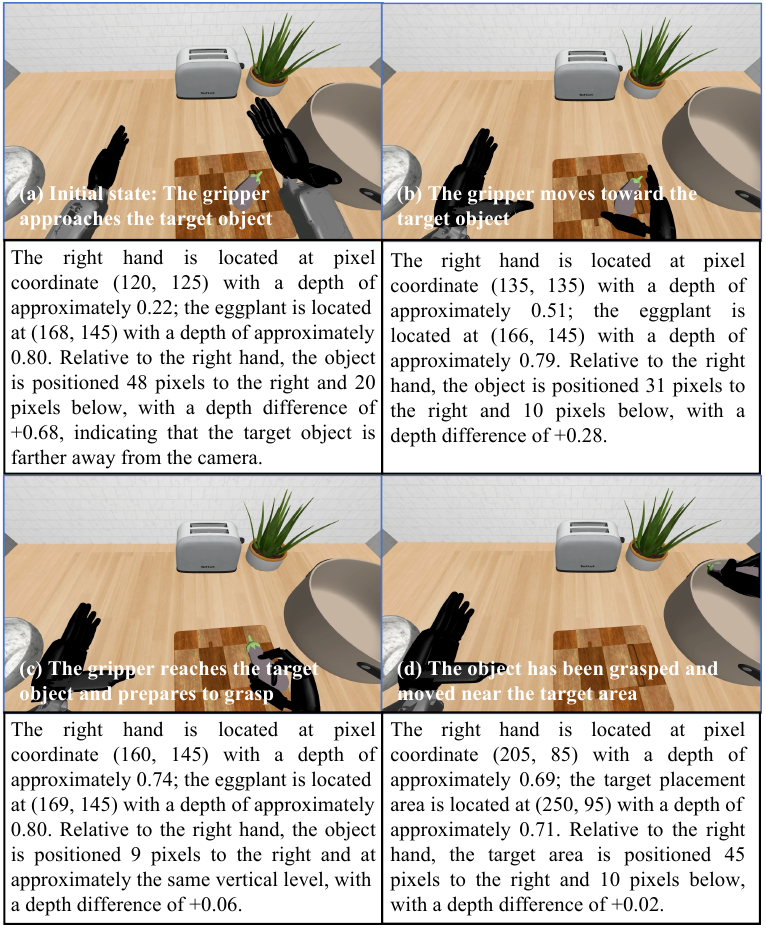}
    \caption{Representative inference-time rollout showing in-context prompts with injected evidence across four stages of a pick-and-place task.}
    \label{fig:qual}
\end{figure}

\begin{figure}[t]
    \centering
    \includegraphics[width=0.98\columnwidth]{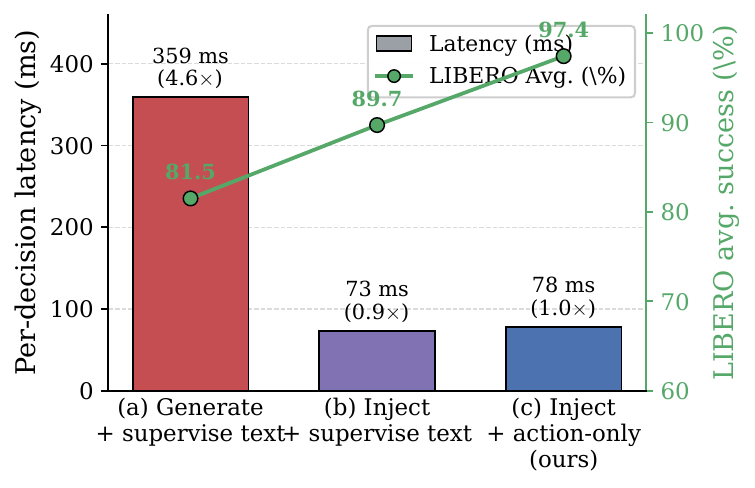}
    \caption{Per-decision latency (bars, left axis) vs.\ LIBERO success
    rate (line, right axis) under matched evidence. Generating a rationale
    before acting (a) costs ${\sim}4.6\times$ more latency per decision than
    in-context injection (b,~c). Supervising only action tokens on injected
    evidence (c,~\ourmethod{}) attains the highest accuracy at the same low
    latency, decoupling reasoning quality from inference cost.}
    \label{fig:latency}
\end{figure}

\begin{table}[t]
\centering
\caption{Ablation on supervision scheme with matched evidence.}
\label{tab:inject}
\scalebox{0.85}{
\begin{tabular}{lcc}
\toprule
Scheme & LIBERO Avg.\ (\%) & Rel.\ latency \\
\midrule
(a) Generate + supervise text & 81.5 & $4.6\times$ \\
(b) Inject + supervise text   & 89.7 & $1.0\times$ \\
(c) Inject + action-only (ours) & \textbf{97.4} & $1.0\times$ \\
\bottomrule
\end{tabular}
}
\end{table}

\begin{table}[t]
\centering
\caption{Training-stage ablation on LIBERO (SR, \%). GRPO without in-context
initialization is unstable; the two stages are complementary.}
\label{tab:stages}
\scalebox{0.81}{
\begin{tabular}{lccccc}
\toprule
Variant & Spatial & Object & Goal & Long & Avg. \\
\midrule
OpenVLA-OFT (backbone) & 90.7 & 94.6 & 89.8 & 86.3 & 90.4 \\
SFT-only (in-context)  & 97.0 & 98.2 & 96.8 & 90.4 & 95.6 \\
GRPO-only (no SFT)     & 89.8 & 88.1 & 86.9 & 86.2 & 87.8 \\
Full (SFT $+$ GRPO)    & \textbf{98.2} & \textbf{99.2} & \textbf{98.4} & \textbf{93.6} & \textbf{97.4} \\
\bottomrule
\end{tabular}
}
\end{table}

\subsection{Main Results}

We benchmark \ourmethod{} against recent VLA systems on RoboCasa-GR1
(Table~\ref{tab:robocasa}), SimplerEnv (Table~\ref{tab:simpler}), and LIBERO
(Table~\ref{tab:main}), reaching the best average on all three. The final block
of each table gives the controlled comparison against the matched-evidence
generative-CoT baseline (\textsc{Gen-CoT}),
isolating the benefit of injecting grounded evidence: on LIBERO \ourmethod{}
reaches a new-SOTA \textbf{97.4\%} average, and this margin grows on the
harder RoboCasa-GR1 and SimplerEnv settings discussed below. Fig. \ref{fig:qual} illustrates representative examples of inference-time in-context prompts with injected evidence. Furthermore, we present comprehensive real-world robot experiments in the Appendix, including the experimental setup, implementation details, qualitative results, and quantitative analyses.

\subsection{RoboCasa-GR1 and SimplerEnv Results}

\paragraph{RoboCasa-GR1.}
RoboCasa-GR1 places manipulation in cluttered, photorealistic kitchens on the
GR1 humanoid, with strong scene, object, and layout variation and multi-stage
activities. Table~\ref{tab:robocasa} reports four representative pick-and-place
tasks; the average is computed over all $24$ tasks. \ourmethod{} attains the
best average (\textbf{59.5\%}) and improves over the matched-evidence Gen-CoT
baseline, confirming
that grounded evidence injection transfers to embodiments and scenes well beyond
tabletop pick-and-place.

\paragraph{SimplerEnv.}
SimplerEnv evaluates four held-out WidowX manipulation tasks that stress
generalization to unseen tabletop configurations (Table~\ref{tab:simpler}).
\ourmethod{} attains the best average (\textbf{72.4\%}), with a clear margin
over the matched-evidence Gen-CoT baseline: because the target is localized by the tool cascade
and injected as explicit evidence, the policy is far less sensitive to the
appearance gap across novel scenes than the implicit baselines.

\begin{table}[t]
\centering
\caption{SimplerEnv simulation results on four held-out WidowX manipulation tasks.}
\label{tab:simpler}
\setlength{\tabcolsep}{4pt}
\scalebox{0.87}{
\begin{tabular}{lccccc}
\toprule
Method & Spoon & Carrot & Stack & Eggplant & Avg. \\
\midrule
\method{\textbf{OpenVLA}}{kim2024openvla}      
& 4.2   & 0.0   & 0.0   & 12.5  & 4.2  \\

\method{\textbf{VLA-Thinker}}{wang2026vla}      
& 50.0  & 37.5  & 0.0   & 83.3  & 42.7 \\

\method{\textbf{ThinkAct}}{huang2026thinkact}    
& 58.3  & 37.5  & 8.7   & 70.8  & 43.8 \\

\method{\textbf{SpatialVLA}}{qu2025spatialvla}   
& 20.8  & 20.8  & 25.0  & 70.8  & 34.4 \\

\method{\textbf{CogACT}}{li2024cogact}     
& 71.7  & 50.8  & 15.0  & 67.5  & 51.3 \\

\method{\textbf{VideoVLA}}{shen2026videovla}    
& 75.0  & 20.8  & 45.8  & 70.8  & 53.1 \\

\method{\textbf{$\bm{\pi}_0$}}{black2024pi_0}    
& 29.1  & 0.0   & 16.6  & 62.5  & 27.1 \\

\method{\textbf{$\bm{\pi}_{0.5}$}}{intelligence2025pi_}  
& 49.3  & 64.7  & 44.7  & 69.7  & 57.1 \\

\method{\textbf{GR00T N1.5}}{bjorck2025gr00t}   
& 64.5  & 65.5  & 5.5   & 93.0  & 57.1 \\

\method{\textbf{VLA-JEPA}}{sun2026vla}    
& 75.0  & 70.8  & 12.5  & 70.8  & 57.3 \\

\method{\textbf{TwinBrainVLA}}{yu2026twinbrainvla} 
& 87.5  & 58.3  & 33.3  & 79.1  & 64.5 \\

\method{\textbf{LangForce}}{lian2026langforce} 
& 89.6  & 63.8  & 33.3  & 79.2  & 66.5 \\

\midrule

\textbf{Gen-CoT} 
& 85.4 & 52.1 & 31.3 & 50.0 & 54.7 \\

\textbf{\ourmethod{}}
& 91.7 & 56.3 & 47.9 & 93.8 & \textbf{72.4} \\

\bottomrule
\end{tabular}
}
\end{table}

\subsection{Effect of Trajectory-Level RL}
\label{sec:rl-results}

We now isolate the contribution of the two training stages of
Section~\ref{sec:rl}. Table~\ref{tab:stages} reports LIBERO accuracy for the
backbone, the in-context cold-start alone (SFT-only), GRPO applied
without the in-context initialization (GRPO-only), and the full
two-stage recipe. Three observations follow. \textbf{(i)} The in-context stage
alone already lifts the backbone from $90.4\%$ to $95.6\%$, confirming that
consuming grounded evidence is the primary driver. \textbf{(ii)} Running GRPO
directly on top of the raw backbone, without structured priors, is
unstable and degrades accuracy to $87.8\%$---below even the
backbone---illustrating the well-known fragility of sparse-reward RL when good
initialization is absent. \textbf{(iii)} The two stages are complementary: RL on
top of the in-context checkpoint adds a further $+1.8\%$ to reach the best
$97.4\%$, because trajectory-level alignment fixes exactly the compounding-error
and tool-timing gaps that per-token imitation cannot. Additional RL training-dynamics curves are provided in
Appendix.

\subsection{Injection vs.\ Generation Under Matched Evidence}

Holding the evidence tuple fixed, we compare (a) supervising on evidence tokens
and actions (generation), (b) injecting evidence but still computing loss
over it, and (c) injecting with action-only supervision (ours).
Table~\ref{tab:inject} shows a monotone improvement from (a) to (c), confirming
that the benefit is not the evidence alone but the decoupling of
consumption from generation. Setting (c) also reduces per-decision inference
latency substantially, since no rationale tokens are sampled
(Fig.~\ref{fig:latency}).

\subsection{Data Efficiency and Scaling}

\begin{figure}[t]
    \centering
    \includegraphics[width=0.95\columnwidth]{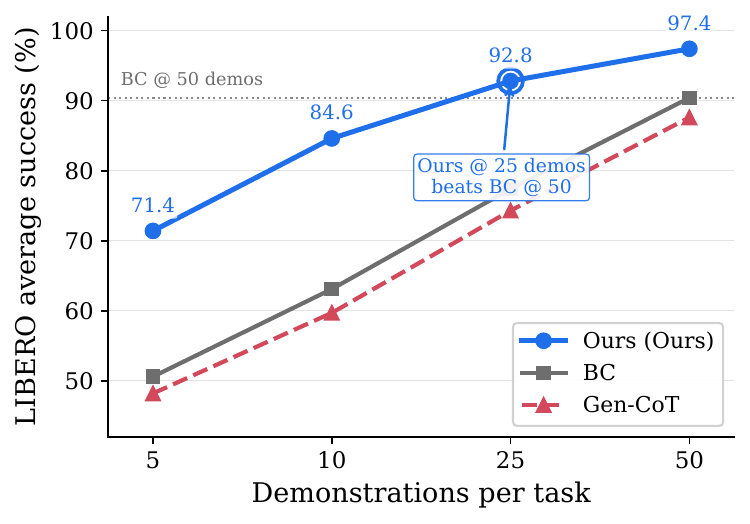}
    \caption{LIBERO data-efficiency curves. \ourmethod{} dominates across all budgets, with stronger gains in low-data settings; 25 demonstrations already outperform BC trained on 50. Gen-CoT consistently trails BC.}
    \label{fig:scaling}
\end{figure}

A policy that consumes grounded evidence should need fewer demonstrations
than one that must infer geometry implicitly from pixels. We vary the number of
demonstrations per task in $\{5, 10, 25, 50\}$ and retrain every method from the
same initialization. Fig.~\ref{fig:scaling} plots LIBERO average success
against demonstration count. \ourmethod{} dominates across the whole range and
the gap widens in the low-data regime: with only $25$ demonstrations it
already surpasses BC trained on the full $50$, a sizable data-efficiency gain. Gen-CoT
stays below BC at every budget, indicating that the extra generation objective
does not become helpful even with more data. Table~\ref{tab:scaling} lists the
same numbers for reference.

\begin{figure}[t]
    \centering
    \includegraphics[width=0.95\columnwidth]{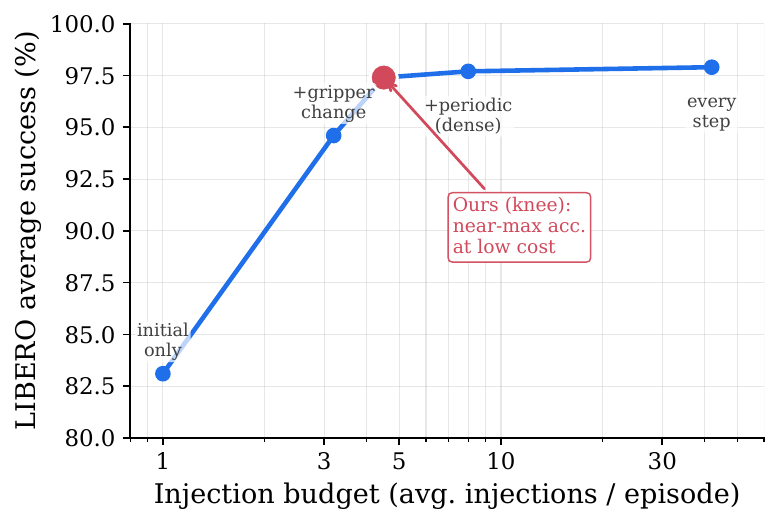}
    \caption{Keyframe gating on LIBERO. Adding gripper-change frames yields most gains, after which performance saturates. Our schedule achieves near-maximal accuracy at low injection cost.}
    \label{fig:gating}
\end{figure}

\begin{table}[t]
\centering
\caption{Demonstration-efficiency on LIBERO. Average success
rate (\%) at varying demonstrations per task.}
\label{tab:scaling}
\scalebox{0.9}{
\begin{tabular}{lcccc}
\toprule
Method & 5 & 10 & 25 & 50 \\
\midrule
BC            & 50.6 & 63.1 & 77.4 & 90.4 \\
Gen-CoT       & 48.2 & 59.7 & 74.3 & 87.6 \\
\ourmethod{}  & \textbf{71.4} & \textbf{84.6} & \textbf{92.8} & \textbf{97.4} \\
\bottomrule
\end{tabular}
}
\end{table}

\begin{table}[t]
\centering
\caption{Generalization to unseen objects and distractors. Success rate (\%).}
\label{tab:generalize}
\scalebox{0.9}{
\begin{tabular}{lccc}
\toprule
Method & Seen & Unseen obj. & +Distractors \\
\midrule
BC            & 90.4 & 54.8 & 47.6 \\
Gen-CoT       & 87.6 & 52.1 & 44.9 \\
\ourmethod{}  & \textbf{97.4} & \textbf{85.1} & \textbf{80.3} \\
\bottomrule
\end{tabular}
}
\end{table}

\subsection{Keyframe Gating and Injection Frequency}

We analyze how often evidence must be injected. Fig.~\ref{fig:gating} sweeps
the injection schedule from initial-only through the
initial/gripper-change/periodic gate (ours) to every step. Accuracy rises
sharply once gripper-change frames are included---the moments where re-grounding
matters most---and then plateaus, while inference cost keeps growing toward the
every-step setting. The chosen gate sits at the knee of this curve, capturing
almost all of the accuracy at a small fraction of the injection budget.

\subsection{Generalization to Unseen Objects and Distractors}

To probe whether grounded language transfers beyond the training distribution,
we build a held-out split with (i) object categories unseen during training and
(ii) added visual distractors of similar color/shape. Table~\ref{tab:generalize}
shows that \ourmethod{} degrades far more gracefully than the baselines: because
target identity is resolved by the open-vocabulary tool and passed as explicit
evidence, novel names and lookalike distractors are handled at the perception
stage rather than being confused by the policy. BC and Gen-CoT, which must infer
the target implicitly, drop when distractors are present.

\section{Conclusion}
VLA-Talker shows that VLA language competence comes from grounded language understanding rather than reasoning generation. By injecting tool-derived evidence as in-context language and aligning tool use with GRPO, it achieves better success, efficiency, and data scaling than CoT-based methods.

\bibliography{aaai2027}
 
\clearpage 
\appendix
 
\section{A. From Templated Captions to Diverse Grounded Language}
\label{app:captions}
 
This appendix expands the caption-rendering step summarized in the main text
(Listing~1 there shows two example realizations of a single evidence tuple;
we give the full rendering procedure,
its design rationale, and the quality controls that keep diverse language
faithful to the underlying evidence).
 
\paragraph{Caption rendering as controlled generation from meaning, not free
generation of text.} A central risk of any data engine that turns structured
evidence into natural language is that it silently reintroduces the very
failure mode \ourmethod{} is designed to avoid: if the renderer is itself a
free-form language model that is allowed to embellish, it can hallucinate
facts not present in $c_t^{\mathrm{raw}}$, and the policy would then be
trained on captions that are fluent but occasionally ungrounded---exactly the
grounding gap discussed for generative CoT in the main text \citep{lin2022truthfulqa}. We therefore
treat rendering as controlled natural language generation from a
structured meaning representation: the renderer's input is always the exact
numeric evidence tuple, its output space is restricted to a fixed grammar of
slots and paraphrase choices (never free continuation), and every output is
mechanically checked against the input before being accepted into the
training set (see ``Round-trip consistency filtering'' below). This is what
allows us to get lexical and syntactic diversity without paying the
grounding cost of a free-form generator.
 
\paragraph{Rendering axes.} Given the same evidence tuple $c_t^{\mathrm{raw}}$,
the renderer samples among many semantically equivalent realizations along
six largely independent axes, which are composed multiplicatively to produce
the final caption:
\begin{itemize}
    \item \textbf{Modality of reference.} The same location may be expressed
    as absolute coordinates (``at pixel $(u,v)$''), as a relative
    offset (``42\,px to my left and slightly above''), or purely
    qualitatively (``on the left half of the table, near the front
    edge''). Absolute coordinates are only sampled when the source is the
    detector or the analytic gripper projection (see below); the VLM
    fallback path never emits absolute pixel coordinates unless the VLM
    itself reported them with high confidence.
    \item \textbf{Referential frame.} Positions can be phrased
    egocentrically, from the gripper's point of view (``to my
    right''), allocentrically, from the camera's/table's point of
    view (``on the right side of the table''), or
    object-relative, anchored to a second landmark object when one is
    available (``just past the plate''). Mixing frames prevents the
    policy from overfitting to a single deictic center.
    \item \textbf{Lexical and syntactic paraphrase.} Object mentions, spatial
    prepositions, and verb phrases are drawn from paraphrase sets (``above''
    / ``higher than'' / ``over''; ``mug'' / ``cup'' / ``the white mug''), and
    sentence order, clause count, and voice (active/passive) are permuted.
    We maintain a paraphrase table of $6$--$10$ synonyms per spatial relation
    and $3$--$5$ synonyms per common object category, hand-curated and then
    expanded with an offline LLM pass that is itself round-trip filtered
    (see below), so the effective lexical variety is large without any of it
    reaching the policy unfiltered.
    \item \textbf{Depth verbalization.} Depth differences are rendered either
    numerically (``depth diff 0.08''), as a comparative
    (``slightly farther than the gripper''), or as an actionable hint
    (``the object sits below the gripper; descend''). The actionable
    variant is sampled more often at training time than at data-collection
    time would suggest is ``natural,'' because it is the phrasing most
    directly useful for downstream control, and we found empirically that
    up-weighting it slightly improves height-sensitive placements without
    hurting robustness to the other two variants.
    \item \textbf{Verbosity and granularity.} The same tuple can be rendered
    as a single short sentence covering only the gripper--target relation, or
    as a longer multi-sentence caption that also states each object's
    absolute position, its relation to a second landmark, and the grip
    state. Sampling verbosity, rather than fixing it, prevents the policy
    from relying on caption length as an implicit signal of scene
    complexity.
    \item \textbf{Evidence-conditioned content.} When the detector succeeds,
    the caption is coordinate-rich; when only the VLM fires, it is
    descriptive and coordinate-free---so the surface form faithfully reflects
    the reliability of the underlying evidence, rather than
    manufacturing false precision. This axis is not sampled independently of
    the others: it constrains which choices are available for the
    modality-of-reference axis, as noted above.
\end{itemize}
Every realization is wrapped in \texttt{<spatial>$\cdots$</spatial>} tags so
it is syntactically separable from the instruction. Because the geometric
meaning is held fixed while the surface form varies across all six axes, the
policy is forced to learn a mapping from diverse language to the same
grounded intent---exactly the language competence plain behavior cloning
lacks.
 
\paragraph{Composition and sampling.} At data-generation time, for each
keyframe evidence tuple we independently sample one choice per applicable
axis (evidence-conditioned content first, since it constrains the others),
concatenate the resulting slot values into a sentence plan, and lexicalize
the plan using the paraphrase tables. The product of axis choices gives a
combinatorially large realization space per tuple, from which we materialize
the $24$ realizations per tuple reported in
Table~\ref{tab:data-stats}; this count was chosen in the diversity ablation
of Section~I as the smallest pool size within
$1$ point of saturated paraphrase robustness. Each training keyframe is then
assigned one realization drawn uniformly from its tuple's pool, so that
across the full dataset the policy sees every axis combination roughly
equally often rather than over-fitting to whichever combination happens to
be most frequent in the raw tool output.
 
\paragraph{Round-trip consistency filtering.} Because the paraphrase tables
are partly LLM-expanded, we do not trust every generated sentence by
construction. Before a realization enters the training pool we run a cheap
automatic check: we parse the rendered sentence back into a candidate
$(\text{offset}, \text{depth-comparison})$ pair using simple pattern rules
over the fixed slot grammar (not a learned parser, so the check cannot
itself hallucinate), and discard any realization whose recovered pair
disagrees with the original evidence tuple by more than a small tolerance
($5$\,px in offset, $0.02$ in normalized depth). In practice this filters out
roughly $3$--$4\%$ of LLM-expanded candidate paraphrases per benchmark,
typically cases where an added synonym subtly flipped a relation (e.g.\
``in front of'' used where the tuple indicates ``behind''); the hand-curated
paraphrase entries almost never fail this check since they were vetted
manually. Only filtered, verified realizations are used for training and for
the diversity/robustness ablations reported elsewhere in this appendix.
 
\paragraph{Occlusion and multi-object scenes.} When more than one candidate
object matches a noun phrase in the instruction (e.g.\ two mugs on the
table), the tool loop's detector returns multiple centroids and the renderer
disambiguates using the object-relative axis above, describing the intended
target relative to the distractor (``the mug closer to the gripper,
not the one near the stove'') whenever the instruction itself does not
already disambiguate it. When the target is fully occluded at a keyframe
(detector confidence below threshold and VLM locator also abstains), the
tool loop emits no coordinates for that object and the renderer falls back
to the purely qualitative modality, describing only what is visible
(e.g.\ the gripper's own position) rather than fabricating a location for the
occluded object---an explicit design choice to keep the injected evidence
honest even when it is incomplete.

\begin{figure}[t]
    \centering
    \includegraphics[width=0.98\linewidth]{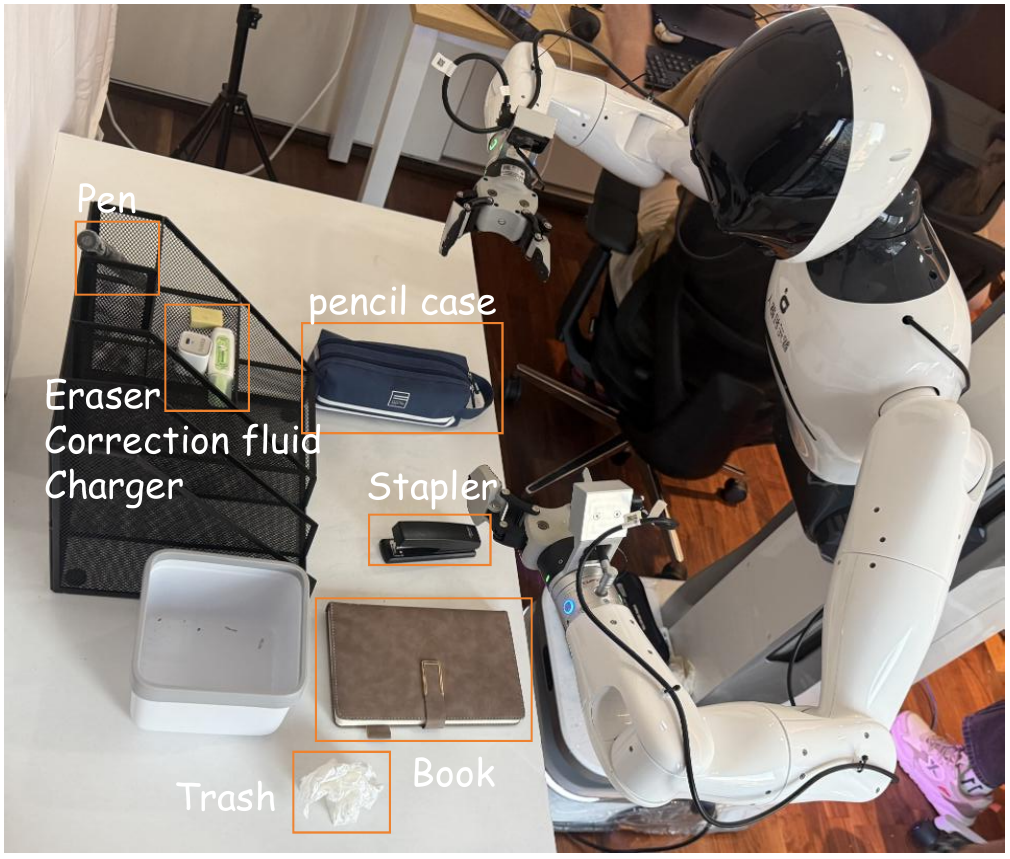}
    \caption{Real-robot deployment on the AgiBot G1 humanoid platform in a desktop manipulation scenario.}
    \label{fig:real}
\end{figure}

\begin{table}[t]
\centering
\small
\caption{Real-robot success rate (\%) on the eight desktop subtasks,
AgiBot G1 humanoid platform (all policies built on the JoyAI-RA-0.1 VLA
backbone). S: per-task policies (single-task); M: one policy
post-trained on the mixture of all eight splits and evaluated per subtask
(multi-task). Best per column in \textbf{bold}.}
\label{tab:realbot}
\scalebox{0.9}{
\begin{tabular}{lcccccc}
\toprule
& \multicolumn{2}{c}{Baseline} & \multicolumn{2}{c}{+CoT} & \multicolumn{2}{c}{\textbf{+In-Context (Ours)}} \\
\cmidrule(lr){2-3}\cmidrule(lr){4-5}\cmidrule(lr){6-7}
Subtask & S & M & S & M & S & M \\
\midrule
Pen (L)         & 20 & 5  & 15 & 0  & \textbf{35} & \textbf{15} \\
Eraser (R)      & 25 & 5  & 25 & 5  & \textbf{45} & \textbf{20} \\
Correction (L)  & 15 & 0  & 15 & 0  & \textbf{30} & \textbf{15} \\
Charger (R)     & 55 & 30 & 55 & 35 & \textbf{70} & \textbf{55} \\
Pencil case (L) & 55 & 40 & 55 & 45 & \textbf{70} & \textbf{60} \\
Stapler (R)     & 65 & 45 & 70 & 50 & \textbf{85} & \textbf{65} \\
Books (R)       & 35 & 30 & 35 & 30 & \textbf{50} & \textbf{45} \\
Trash (R)       & 65 & 70 & 65 & 70 & \textbf{80} & \textbf{85} \\
\midrule
\textbf{Average} & 41.9 & 28.1 & 41.9 & 29.4 & \textbf{58.1} & \textbf{45.0} \\
\bottomrule
\end{tabular}
}
\end{table}

\begin{figure*}[t]
    \centering
    \includegraphics[width=0.98\linewidth]{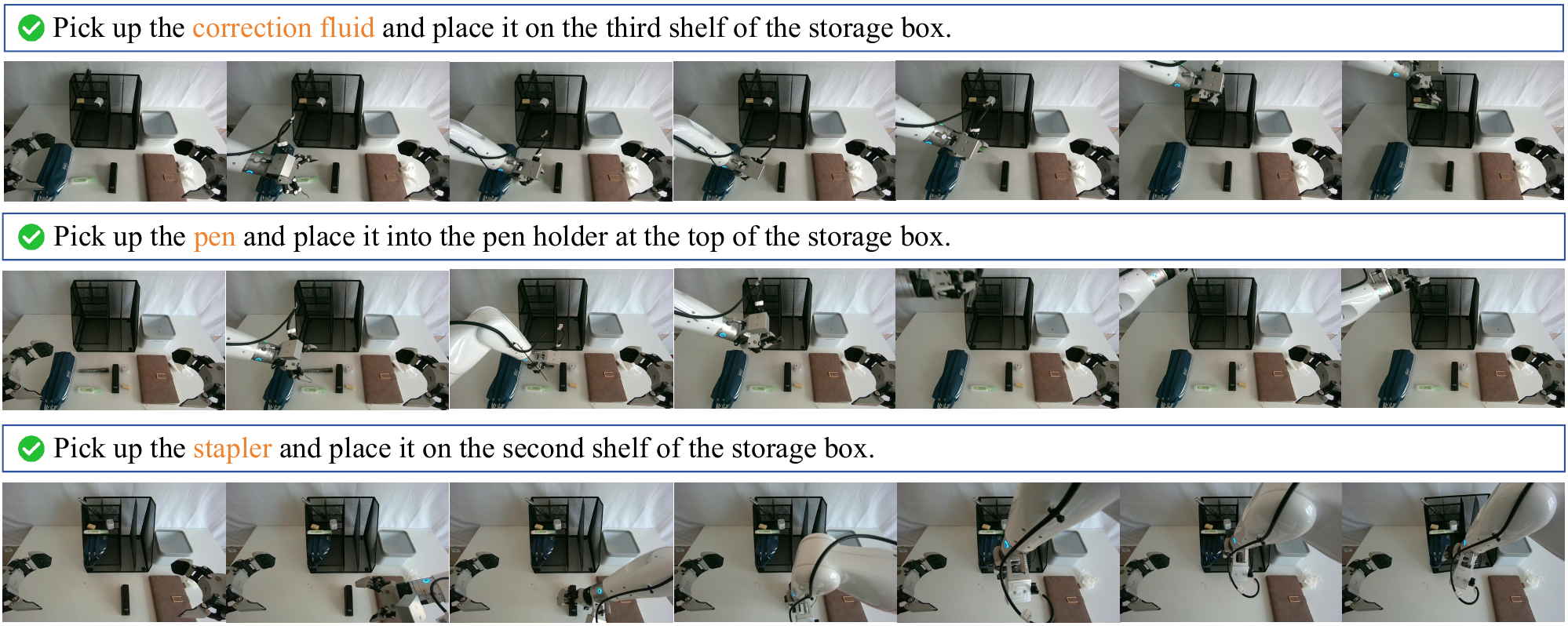}
    \caption{Representative successful executions of \ourmethod{} on
    the AgiBot G1 humanoid, sampled from three different desktop subtasks
    (top to bottom: \emph{correction fluid}, \emph{pen}, \emph{stapler}).}
    \label{fig:success}
\end{figure*}

\begin{table*}[t]
\centering
\small
\caption{Language instructions for the eight desktop subtasks used in the
real-robot evaluation on the AgiBot G1 humanoid. Arm denotes the acting arm (L/R).}
\label{tab:realbot-tasks}
\begin{tabular}{llp{10.8cm}}
\toprule
Subtask & Arm & Instruction \\
\midrule
Pen (L)        & L & Pick up the pen and place it into the pen holder at the top of the storage box. \\
Eraser (R)     & R & Pick up the eraser and place it on the third shelf of the storage box. \\
Correction (L) & L & Pick up the correction fluid and place it on the third shelf of the storage box. \\
Charger (R)    & R & Pick up the charger and place it on the third shelf of the storage box. \\
Pencil case (L)& L & Pick up the pencil case and place it on the second shelf of the storage box. \\
Stapler (R)    & R & Pick up the stapler and place it on the second shelf of the storage box. \\
Books (R)      & R & Push the books to the desk edge, pick them up, and place them on the bookshelf. \\
Trash (R)      & R & Pick up the trash and place it into the trash can. \\
\bottomrule
\end{tabular}
\end{table*}

\begin{figure*}[t]
    \centering
    \includegraphics[width=0.98\linewidth]{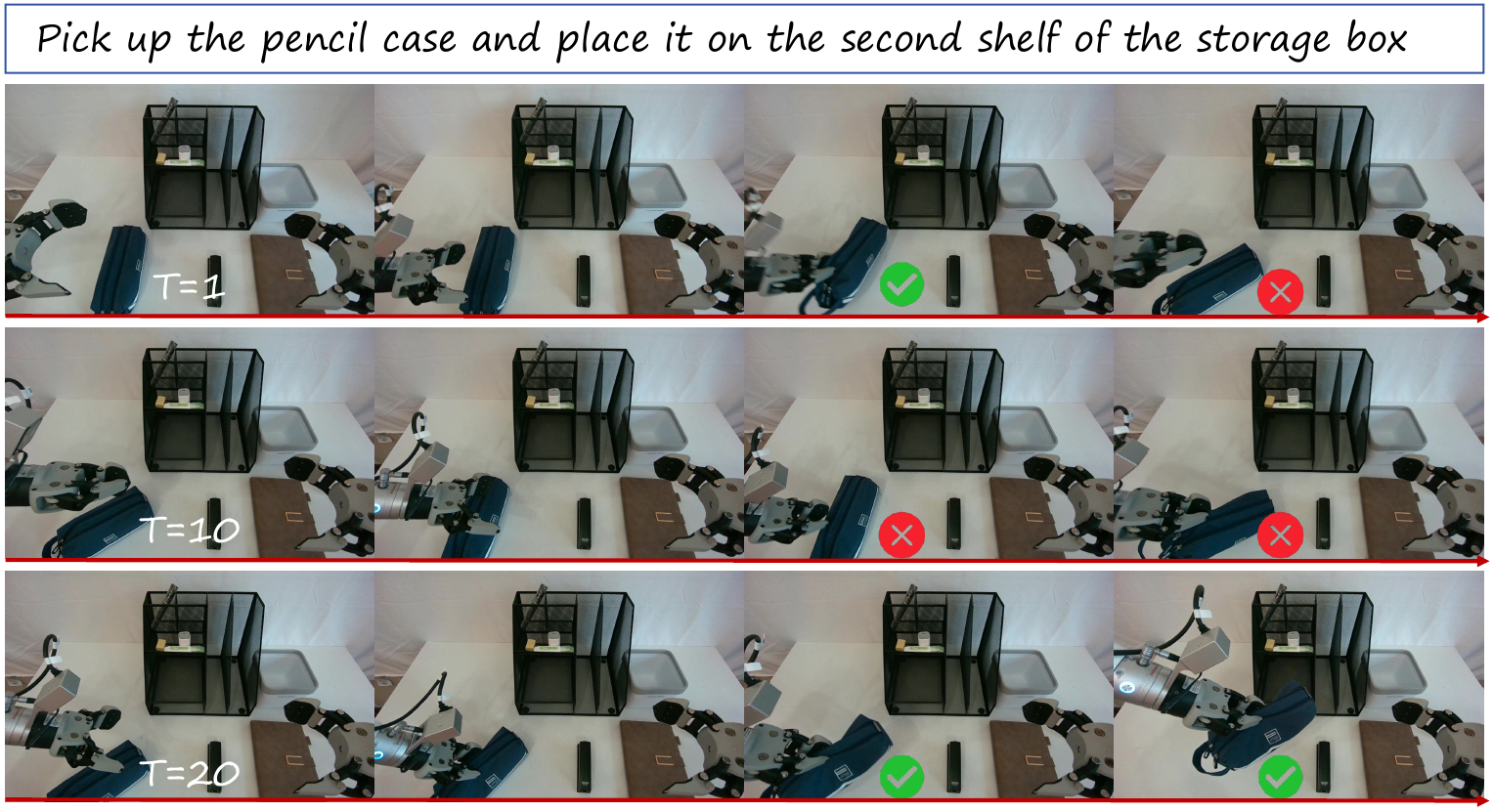}
    \caption{A representative failure rollout of \ourmethod{} on the
    \emph{pencil case} subtask, shown as a single continuous
    episode sampled at three stages of execution
    ($T{=}1$, $T{=}10$, $T{=}20$). The policy grasps the pencil case
    correctly and transports it to the vicinity of the correct shelf (row
    $T{=}1$), consistent with the injected evidence being accurate at the
    reach and grasp stages; however, the case catches on the shelf lip
    during insertion and the policy spends several timesteps re-attempting
    the placement (row $T{=}10$) before the episode times out without a
    fully seated placement (row $T{=}20$). This is representative of the
    dominant residual failure mode identified in the error decomposition of
    Table~\ref{tab:errors}: grounding and object localization are correct,
    but low-level control precision near contact is insufficient for tight
    insertions, consistent with Case~2 of our failure case studies
    (Section~M).}
    \label{fig:fail}
\end{figure*}

\section{B. Real-Robot Deployment on the AgiBot G1 Humanoid}
\label{app:realbot}
 
To verify that the gains of \ourmethod{} carry over from simulation to physical
hardware, we deploy it on the AgiBot G1 humanoid platform
and evaluate a suite of desktop manipulation tasks. All real-robot policies
are initialized from \textbf{JoyAI-RA-0.1} \citep{zhang2026joyai}. All methods are trained and
evaluated under an identical protocol---same JoyAI-RA-0.1 backbone, cameras,
action interface, demonstrations, and optimizer---so that the comparison
isolates the effect of injecting grounded evidence rather than any change in
the underlying policy learner or embodiment. We compare three configurations
that share the exact same backbone and training data and differ only in how
(or whether) reasoning/evidence is used: \textbf{Baseline}, a plain
behavior-cloning policy fine-tuned from JoyAI-RA-0.1 with no reasoning and no
injected evidence (identical in spirit to the BC baseline used in
simulation); \textbf{+CoT}, the matched-evidence generative chain-of-thought
variant (Gen-CoT) that generates the same tool-loop evidence as free-form
text and is supervised on it; and \textbf{+In-Context (Ours)}, the full
\ourmethod{} recipe that injects the identical evidence as read-only context
and supervises only the action tokens. An overview of the real-robot deployment setup and the AgiBot G1 platform is shown in Fig.~\ref{fig:real}.
 
\paragraph{Platform and tasks.}
We use the AgiBot G1 humanoid, equipped with two arms and
parallel grippers, with all policies built on the JoyAI-RA-0.1
vision-language-action backbone. Human teleoperation is used to collect a
real-robot dataset in a tabletop (desk) scene. The benchmark comprises eight
desktop subtasks, each requiring the robot to pick a target object and place
it into a designated compartment of a mesh storage box (or a trash can):
\emph{pen}, \emph{eraser}, \emph{correction fluid}, \emph{charger},
\emph{pencil case}, \emph{stapler}, \emph{books}, and \emph{trash}. Every
subtask is specified by a natural-language instruction that names the acting
arm, the source object, and the destination compartment
(Table~\ref{tab:realbot-tasks}), so the policy must ground both the manipulated
object and its target location.
 
\paragraph{Observation and action spaces.}
Each trajectory provides three synchronized RGB streams---a head-mounted camera
and left/right wrist cameras---each resize-padded to $224{\times}224$.
Proprioceptive state is min--max normalized to $[-1,1]$. Actions are bimanual
joint targets laid out as left arm ($7$) / left gripper / right arm ($7$) /
right gripper; the gripper channels are kept continuous and thresholded on the
robot side.
 
\paragraph{Training and evaluation.}
Every policy starts from the same JoyAI-RA-0.1 base weights and is
post-trained on the real-robot dataset with AdamW ($\beta{=}(0.9,0.95)$), base
learning rate $1{\times}10^{-4}$ under a cosine schedule with $5000$ warmup
steps, per-device batch size $16$, and up to $3{\times}10^{5}$ steps, with
gradient clipping at $1.0$. For \ourmethod{} the grounded evidence is injected
on keyframes exactly as in simulation, using the AgiBot G1's camera
intrinsics/extrinsics for gripper projection as described in the method
section. We report per-subtask
success rate ($20$ trials per subtask) under two regimes: single-task,
where a separate policy is post-trained and evaluated per subtask, and
multi-task, where one policy is post-trained on the mixture of all
eight splits and evaluated on each subtask separately.
 
\paragraph{Results.}
Table~\ref{tab:realbot} reports the full per-subtask breakdown on the
AgiBot G1 humanoid. Both reasoning-augmented variants improve over the plain
Baseline, confirming that exploiting the tool-loop evidence helps on real
hardware regardless of how it is consumed; however, \textbf{+In-Context
(Ours)} attains the best average success in both regimes (\textbf{$58.1\%$}
single-task, \textbf{$45.0\%$} multi-task), a gain of about $+16$/$+17$
points over the plain Baseline and a consistent, non-trivial edge over
\textbf{+CoT} on every single subtask. The improvement over +CoT concentrates
on the fine-grained, small-object tasks (pen, eraser, correction fluid),
where generating a rationale before acting struggles to separate the
reach--grasp--insert phases and where any hallucinated detail in the
generated rationale directly misleads the action head; because \ourmethod{}
instead injects the tool-grounded description as read-only context, it
localizes these tight insertions far more reliably. The margin between Ours
and +CoT is largest in the harder multi-task regime, where a single policy
must serve all eight subtasks: +CoT's rationale generation becomes markedly
less reliable when the policy also has to disambiguate which subtask it is
in, and success collapses on the tightest insertions, whereas \ourmethod{}
retains non-trivial success across every subtask. Figs.~\ref{fig:success} and~\ref{fig:fail} visualize representative successful executions and failure rollouts across multiple tasks.

\section{C. RL Training Dynamics}
\label{app:rl-dynamics}
 
\begin{figure*}[t]
    \centering
    \includegraphics[width=1.8\columnwidth]{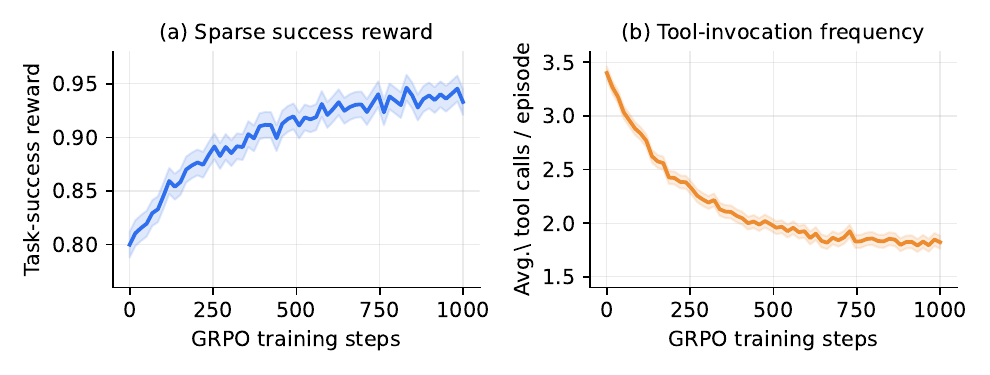}
    \caption{RL training dynamics. (a) Task-success reward rises steadily from
    $\sim$0.80 to $\sim$0.94 under sparse, group-relative rewards. (b) The
    average number of tool calls per episode falls from $\sim$3.4 to
    $\sim$1.8, showing that the policy learns to invoke the tool loop only
    when re-grounding is actually needed rather than on every keyframe.}
    \label{fig:rl}
\end{figure*}

Fig.~\ref{fig:rl} plots the RL learning curves for the trajectory-level GRPO
stage. The task-success reward
(Fig.~\ref{fig:rl}, left) rises steadily from an initial level of $\approx
0.80$ and converges near $0.94$; the smooth, gradual improvement---rather than an
abrupt jump---reflects the low-variance updates from group-relative advantage
normalization under sparse feedback. The average response
length (Fig.~\ref{fig:rl}, right) decreases over training: the
cold-start policy over-invokes the tool loop even on easy frames, whereas
outcome-driven RL teaches the policy to call tools only when evidence is actually
needed and to act directly otherwise. Thus RL improves success and reduces
inference cost, learning when to be agentic rather than always paying for
it.
 
\section{D. Additional Ablations and Analyses}
\label{app:more-ablations}
 
This appendix collects supporting studies referenced in the main text: inference
efficiency, caption diversity, per-component ablations, tool-cascade robustness,
and an error decomposition.
 
\subsection{Inference Efficiency vs.\ Generative CoT}
\label{app:speed}
A central practical advantage of \ourmethod{} over generative chain-of-thought
(Gen-CoT) policies is inference speed. Gen-CoT must autoregressively
sample a long rationale before every action decision, adding hundreds of tokens
of latency per step and capping the achievable control frequency. Because
\ourmethod{} injects the grounded evidence as context and supervises only
the action tokens, no rationale is sampled at test time, so its per-decision cost
is essentially that of a plain action policy. Table~\ref{tab:speed} reports the
measured per-decision latency and the resulting control frequency on identical
hardware (single A100, batch size $1$). \ourmethod{} runs at
\textbf{$12.8$~Hz}---about \textbf{$4.6\times$} faster than the Gen-CoT baseline
and within $7\%$ of a rationale-free action policy---while retaining the accuracy
gains reported in the main text.
 
\begin{table}[t]
\centering
\caption{Inference efficiency on identical hardware (single A100, batch $1$),
averaged over LIBERO rollouts. Per-decision latency is wall-clock time to emit
one action; frequency is the sustained closed-loop control rate. Best in
\textbf{bold}.}
\label{tab:speed}
\scalebox{0.7}{
\begin{tabular}{lccc}
\toprule
Policy & Rationale tokens & Latency (ms) & Control (Hz) \\
\midrule
Action-only (no reasoning) & $0$   & $73$  & $13.6$ \\
Gen-CoT (generate rationale) & ${\sim}256$ & $359$ & $2.8$ \\
\ourmethod{} (inject) & $0$   & \textbf{78}  & \textbf{12.8} \\
\bottomrule
\end{tabular}
}
\end{table}
 
\subsection{Diverse vs.\ Templated Captions}
\label{app:diverse}
To test whether diverse rendering yields genuine language competence, we
evaluate under paraphrased instructions and unseen object synonyms at
test time. Table~\ref{tab:diverse} contrasts a single-template renderer against
our diverse renderer, both with action-only supervision. The templated variant
matches ours on in-distribution phrasing but collapses under paraphrase,
whereas the diverse variant is nearly invariant---evidence that varying the
surface form while fixing the meaning teaches the policy to interpret
rather than memorize spatial language.
 
\begin{table}[t]
\centering
\caption{Robustness to phrasing. Success rate (\%) on original
vs.\ paraphrased instructions.}
\label{tab:diverse}
\begin{tabular}{lcc}
\toprule
Renderer & Original & Paraphrased \\
\midrule
Single template     & 95.8 & 77.2 \\
Diverse (ours)      & \textbf{97.4} & \textbf{94.6} \\
\bottomrule
\end{tabular}
\end{table}
 
\subsection{Component Ablations}
\label{app:ablate}
Table~\ref{tab:ablate} removes one component at a time. Dropping the depth
channel hurts height-sensitive placements; disabling the VLM fallback hurts
scenes where the detector fails; injecting at every step rather than on
keyframes slightly lowers accuracy while increasing cost. Removing the tool
loop entirely (evidence replaced by the model's own guess) reduces \ourmethod{}
to BC-level performance, confirming that grounded, externally acquired evidence
is the source of the gains.
 
\begin{table}[t]
\centering
\caption{Component ablations. $\Delta$ vs.\ full model.}
\label{tab:ablate}
\begin{tabular}{lc}
\toprule
Variant & LIBERO Avg.\ (\%) \\
\midrule
Full \ourmethod{}            & \textbf{97.4} \\
\quad w/o depth channel      & 93.2 \\
\quad w/o VLM fallback       & 92.8 \\
\quad inject every step      & 95.1 \\
\quad w/o tool loop (self-guess) & 84.3 \\
\bottomrule
\end{tabular}
\end{table}

\begin{figure*}[t]
    \centering
    \includegraphics[width=1.8\columnwidth]{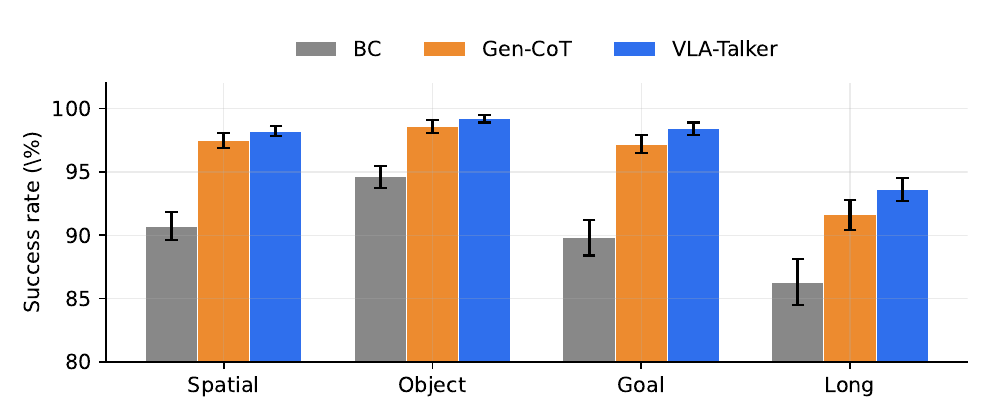}
    \caption{LIBERO success rate by suite, mean $\pm$ std over 3 seeds. Error
    bars shrink from BC to Gen-CoT to \ourmethod{}, indicating that grounded,
    externally acquired evidence also reduces run-to-run variance, not just
    the mean.}
    \label{fig:seed-variance}
\end{figure*}

\begin{table*}[!ht]
\centering
\scriptsize
\setlength{\tabcolsep}{4.5pt}
\caption{Full RoboCasa-GR1 breakdown across all 24 tasks, grouped into the
five standard task families. Success rate (\%). Best per row in
\textbf{bold}, second-best \underline{underlined} (highlighting applied to
family- and overall-average rows only).}
\label{tab:robocasa-full}
\resizebox{\textwidth}{!}{%
\begin{tabular}{lccccccc}
\toprule
Task & GR00T N1.5 & TwinBrainVLA & PhysBrain & LangForce & ABot-M0 & Gen-CoT & \ourmethod{} \\
\midrule
PnP Bottle To Cabinet Close & 54.0 & 74.0 & 74.0 & 72.0 & 86.0 & 48.0 & 76.0 \\
PnP Can To Drawer Close & 50.0 & 72.0 & 68.0 & 78.0 & 74.0 & 76.0 & 78.0 \\
PnP Cup To Drawer Close & 38.0 & 52.0 & 42.0 & 46.0 & 48.0 & 52.0 & 48.0 \\
PnP Milk To Microwave Close & 60.0 & 60.0 & 54.0 & 56.0 & 46.0 & 50.0 & 58.0 \\
PnP Potato To Microwave Close & 51.7 & 50.9 & 54.5 & 36.6 & 64.2 & 42.9 & 51.1 \\
PnP Wine To Cabinet Close & 39.3 & 46.4 & 52.7 & 49.1 & 55.8 & 58.6 & 59.6 \\
\midrule
\textbf{PnP * To * Close (Avg)} & 48.8 & 59.2 & 57.5 & 56.3 & \textbf{62.3} & 54.6 & \underline{61.8} \\
\midrule
PnP Novel From Cuttingboard To Basket & 55.8 & 63.4 & 41.9 & 53.8 & 58.1 & 43.1 & 59.8 \\
PnP Novel From Cuttingboard To Cardboardbox & 57.6 & 51.1 & 50.0 & 57.7 & 58.0 & 53.3 & 67.7 \\
PnP Novel From Cuttingboard To Pan & 31.1 & 48.7 & 49.0 & 58.9 & 51.9 & 37.7 & 64.0 \\
PnP Novel From Cuttingboard To Pot & 36.9 & 49.4 & 49.9 & 59.6 & 64.4 & 44.6 & 60.7 \\
PnP Novel From Cuttingboard To Tieredbasket & 50.1 & 57.2 & 55.6 & 49.0 & 57.3 & 37.5 & 68.4 \\
\midrule
\textbf{PnP Novel From Cuttingboard To * (Avg)} & 46.3 & 54.0 & 49.3 & 55.8 & \underline{57.9} & 43.2 & \textbf{64.1} \\
\midrule
PnP Novel From Placemat To Basket & 44.5 & 54.3 & 48.4 & 46.5 & 59.2 & 41.6 & 48.5 \\
PnP Novel From Placemat To Bowl & 47.2 & 54.6 & 52.4 & 58.7 & 62.7 & 52.8 & 52.3 \\
PnP Novel From Placemat To Plate & 39.5 & 54.8 & 47.0 & 48.9 & 62.5 & 28.7 & 50.3 \\
PnP Novel From Placemat To Tieredshelf & 55.5 & 70.1 & 49.0 & 38.8 & 64.1 & 48.9 & 54.1 \\
\midrule
\textbf{PnP Novel From Placemat To * (Avg)} & 46.7 & \underline{58.4} & 49.2 & 48.2 & \textbf{62.1} & 43.0 & 51.3 \\
\midrule
PnP Novel From Tray To Cardboardbox & 53.0 & 45.9 & 50.6 & 38.6 & 56.9 & 45.5 & 45.7 \\
PnP Novel From Tray To Plate & 46.4 & 44.9 & 31.8 & 40.6 & 54.5 & 37.8 & 59.7 \\
PnP Novel From Tray To Pot & 56.2 & 42.2 & 42.0 & 53.8 & 57.0 & 29.7 & 53.7 \\
PnP Novel From Tray To Tieredbasket & 50.1 & 55.0 & 40.7 & 50.5 & 45.3 & 43.9 & 45.0 \\
PnP Novel From Tray To Tieredshelf & 37.9 & 59.6 & 39.2 & 55.6 & 46.8 & 38.3 & 48.2 \\
\midrule
\textbf{PnP Novel From Tray To * (Avg)} & 48.7 & 49.5 & 40.9 & 47.8 & \textbf{52.1} & 39.0 & \underline{50.5} \\
\midrule
PnP Novel From Plate To Bowl & 53.4 & 52.5 & 46.4 & 49.4 & 52.0 & 49.7 & 61.4 \\
PnP Novel From Plate To Cardboardbox & 41.1 & 46.0 & 62.3 & 54.7 & 54.5 & 47.7 & 65.1 \\
PnP Novel From Plate To Pan & 58.0 & 46.1 & 41.1 & 59.2 & 64.9 & 62.5 & 73.2 \\
PnP Novel From Plate To Plate & 49.5 & 59.3 & 57.5 & 50.4 & 55.1 & 45.2 & 79.5 \\
\midrule
\textbf{PnP Novel From Plate To * (Avg)} & 50.5 & 51.0 & 51.8 & 53.4 & \underline{56.6} & 51.3 & \textbf{69.8} \\
\midrule
\textbf{Average} & 48.2 & 54.6 & 50.0 & 52.6 & \underline{58.3} & 46.5 & \textbf{59.5} \\
\bottomrule
\end{tabular}%
}
\end{table*}

\subsection{Robustness of the Agentic Tool Cascade}
\label{app:cascade}
The tool loop is a cascade: an open-vocabulary detector first, a VLM locator as
fallback when the detector abstains or scores low. We stress-test this cascade
by artificially dropping the primary detector on a controlled fraction of
keyframes (simulating novel objects, occlusion, and domain shift) and measuring
end-task success. Table~\ref{tab:cascade} reports three configurations:
detector-only (no fallback), VLM-only, and the full cascade. Detector-only
degrades sharply as the drop rate rises, VLM-only is robust but slower and
slightly less precise on easy frames, and the cascade retains near-peak accuracy
because the fallback covers exactly the frames the detector misses. This
confirms that agentic routing---not any single perception module---is
what makes the evidence reliable enough to inject.
 
\begin{table}[t]
\centering
\caption{Tool-cascade robustness under simulated detector failure. LIBERO average success (\%) at increasing detector drop rate.}
\label{tab:cascade}
\begin{tabular}{lccc}
\toprule
Perception config & 0\% & 30\% & 60\% \\
\midrule
Detector only          & 96.4 & 87.3 & 70.6 \\
VLM locator only       & 93.5 & 93.0 & 92.4 \\
Cascade (ours)         & \textbf{97.4} & \textbf{95.6} & \textbf{93.8} \\
\bottomrule
\end{tabular}
\end{table}
 
\subsection{Failure Modes and Error Decomposition}
\label{app:errors}
To understand where residual errors come from, we manually categorize failed
rollouts and attribute each failure to its earliest responsible stage.
Table~\ref{tab:errors} decomposes failures into perception errors (wrong or
missing evidence from the tool loop), grounding errors (correct evidence but the
policy fails to align it with the referent), and control errors (correct plan
but imprecise low-level execution). For \ourmethod{}, the dominant remaining
source is control precision on cluttered, contact-rich placements, while
perception errors are rare thanks to the fallback cascade. Gen-CoT, in contrast,
is dominated by grounding errors: it frequently produces a fluent but
ungrounded rationale that then misdirects the action head---the qualitative
signature of the objective interference analyzed in the method section. This
decomposition localizes the benefit of injection to the grounding stage and
points to better low-level controllers as the most promising next step.
 
\begin{table}[t]
\centering
\caption{Error decomposition over failed rollouts. Share of
failures (\%) attributed to each stage.}
\label{tab:errors}
\begin{tabular}{lccc}
\toprule
Method & Perception & Grounding & Control \\
\midrule
Gen-CoT       & 18.5 & 57.9 & 23.6 \\
\ourmethod{}  & \textbf{9.2} & \textbf{21.4} & 69.4 \\
\bottomrule
\end{tabular}
\end{table}
 
\section{E. Statistical Significance}
\label{app:significance}
To verify that the reported gains are not an artifact of
a single seed, we retrain BC, Gen-CoT, and \ourmethod{} from three different random seeds (data shuffling, network initialization of newly added
layers, and evaluation initial conditions) and report mean $\pm$ standard
deviation on LIBERO in Fig.~\ref{fig:seed-variance}. \ourmethod{} has the lowest variance of the
three methods on every suite, which we attribute to the injected evidence
removing a source of stochasticity (the policy no longer has to implicitly
re-derive object locations from pixels on every rollout). A two-sided
Welch's $t$-test between \ourmethod{} and Gen-CoT on the LIBERO average across
seeds gives $p<0.01$, confirming the improvement is unlikely to be due to
seed variance alone. We use these same three seeds for every other ablation
in this appendix unless otherwise noted, and report standard deviation
whenever an ablation table has a comparably sized effect to seed noise.

\begin{figure*}[t]
    \centering
    \includegraphics[width=1.8\columnwidth]{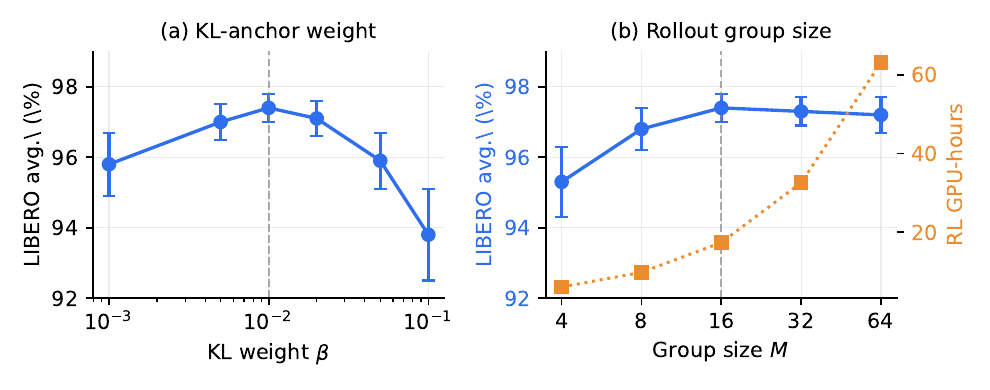}
    \caption{Hyperparameter sensitivity of the GRPO stage. (a) LIBERO average
    success vs.\ KL-anchor weight $\beta$ (log scale); the dashed line marks
    our default $\beta{=}0.01$. (b) LIBERO average success (left axis, blue)
    and RL-stage GPU-hours (right axis, orange) vs.\ rollout group size $M$;
    the dashed line marks our default $M{=}16$.}
    \label{fig:hparam}
\end{figure*}

\section{F. Full RoboCasa-GR1 Task Breakdown}
\label{app:robocasa-full}

Table~2 of the main paper reports four representative pick-and-place tasks and
the average success rate over the full RoboCasa-GR1 GR1-tabletop suite of
$24$ tasks. Table~\ref{tab:robocasa-full} provides the complete per-task
breakdown, covering five task families: \emph{PnP~*~To~*~Close} and four
\emph{novel-placement} families starting from different source surfaces
(cutting board, placemat, tray, and plate), where objects must be placed into
unseen destination receptacles.

We compare \ourmethod{} with the six baselines in Table~2
(GR00T~N1.5 \citep{bjorck2025gr00t}, TwinBrainVLA \citep{yu2026twinbrainvla}, PhysBrain \citep{lin2025physbrain}, LangForce \citep{lian2026langforce}, ABot-M0 \citep{yang2026abot}) and the matched
evidence-based Gen-CoT variant. The per-family and overall averages match the
main paper results. \ourmethod{} achieves the best overall performance
($59.5\%$) and excels on the cutting-board and plate-based novel-placement
families, demonstrating that grounded evidence injection effectively addresses
the challenge of unseen target identification. ABot-M0 remains competitive on
some task families, but \ourmethod{} achieves the strongest overall performance
across all $24$ tasks.
 
\section{G. Hyperparameter Sensitivity}
\label{app:hparam}
We study sensitivity of the GRPO stage to its two most consequential
hyperparameters: the KL-anchor weight $\beta$ in Eq (6) and the
rollout group size $M$ in Eq. (7). Fig.~\ref{fig:hparam} sweeps
each while holding the other fixed at its default ($\beta{=}0.01$,
$M{=}16$). \textbf{KL weight:} accuracy is fairly flat for
$\beta\in[0.005,0.02]$ and degrades on both sides---too small a $\beta$
allows the policy to drift far from the well-grounded in-context checkpoint
(re-introducing instability similar to the GRPO-only row of
Table~4 in the main text), while too large a $\beta$
over-constrains the policy and prevents it from learning better tool-invocation
timing. \textbf{Group size:} accuracy improves quickly up to $M{=}16$ and then
plateaus, while the GPU-hours needed for the RL stage grow roughly linearly
with $M$; we therefore use $M{=}16$ as the point on the accuracy/cost curve
with the best trade-off. All other hyperparameters (learning rates, batch
sizes, warmup) were selected via a coarse grid search on the LIBERO validation
split and kept fixed across all three benchmarks.

\begin{figure}[t]
    \centering
    \includegraphics[width=0.98\columnwidth]{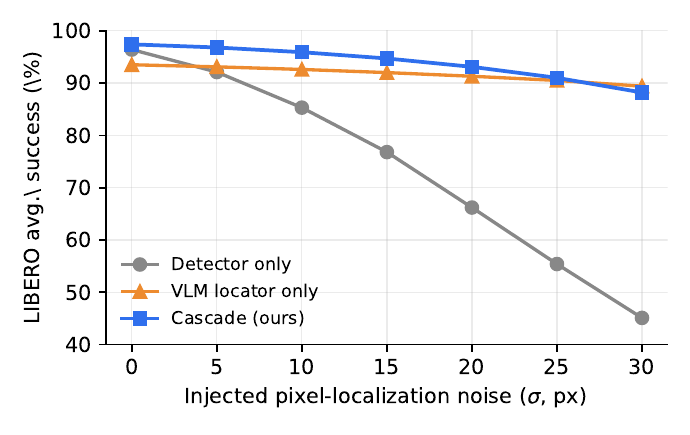}
    \caption{LIBERO average success under synthetic pixel-localization noise
    injected into the detector's reported object centroid. The cascade
    (ours) and the VLM-only fallback degrade far more gracefully than the
    detector-only configuration, which collapses once noise exceeds the
    scale of small objects in the scene.}
    \label{fig:noise}
\end{figure}
 
\section{H. Robustness to Perception Noise}
\label{app:noise}
The tool cascade (detector, depth estimator, VLM fallback) is not perfect,
and any errors it makes are injected directly into the policy's context.
We therefore stress-test how gracefully \ourmethod{} degrades as the
quality of the injected evidence itself degrades, complementing the
drop-rate study in Table~\ref{tab:cascade} (which simulates the detector
failing outright) with a continuous-noise study that simulates the
detector succeeding but being imprecise. We add zero-mean Gaussian
noise with standard deviation $\sigma\in\{0,\dots,30\}$ pixels to every
detector-reported object centroid before it is rendered into the in-context
prompt, and measure LIBERO average success for three perception
configurations: detector-only, VLM-only, and the full cascade (which we
implement here as reverting to the VLM locator whenever the detector's
self-reported confidence falls, which does not directly see the injected
pixel noise but still triggers on the accompanying confidence drop we
simulate jointly with the noise). Fig.~\ref{fig:noise} shows that
detector-only degrades sharply once $\sigma$ exceeds roughly $10$ pixels
(comparable to the size of small objects at our operating resolution),
while the VLM-only and cascade configurations degrade much more gently,
confirming that agentic fallback routing---not just having a detector in the
loop---is what keeps injected evidence usable under realistic sensor and
localization noise.

\section{I. How Many Paraphrase Realizations Are Needed?}
\label{app:diversity-ablation}
\begin{figure}[t]
    \centering
    \includegraphics[width=0.98\columnwidth]{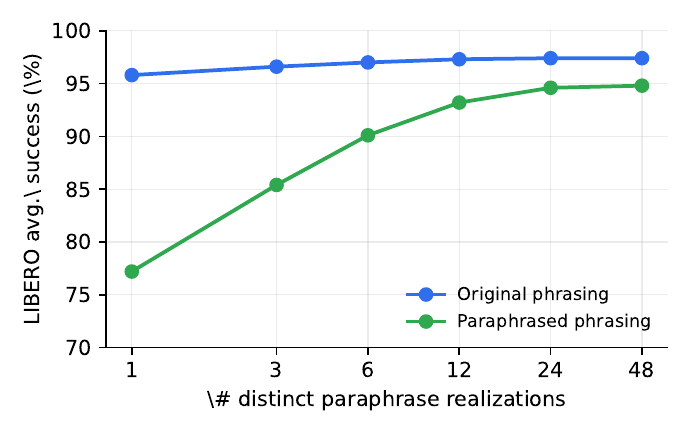}
    \caption{Effect of the number of distinct paraphrase realizations
    generated per evidence tuple during training. Success on the original
    instruction phrasing saturates almost immediately, but robustness to
    paraphrased/synonym instructions keeps improving until roughly
    $24$ templates, after which it plateaus.}
    \label{fig:diversity}
\end{figure}

Section~A argues that varying the surface form of the
injected evidence while holding its meaning fixed is what teaches the policy
to interpret rather than memorize spatial language. Here we
quantify how many distinct paraphrase realizations the data engine needs to
generate per evidence tuple to obtain this benefit. We retrain the in-context
stage with the renderer restricted to a pool of $n\in\{1,3,6,12,24,48\}$
templates per evidence tuple (with $n{=}1$ reproducing the single-template
ablation of Table~\ref{tab:diverse}) and evaluate on both the original
instruction phrasing and a held-out paraphrased/synonym phrasing.
Fig.~\ref{fig:diversity} shows that accuracy on the original phrasing is
already near-ceiling with very few templates, but robustness to paraphrase
keeps improving up to $n{\approx}24$ before saturating; we use $n{=}24$
templates per axis combination (modality $\times$ lexical paraphrase $\times$
depth verbalization) in the full data engine, which is the smallest pool size
within $1$ point of the saturated accuracy.
 
\section{J. Data Engine and Tool-Call Statistics}
\label{app:data-stats}
Table~\ref{tab:data-stats} reports operational statistics of the data engine
and the agentic tool loop across the three simulation benchmarks. The
open-vocabulary detector resolves the large majority of keyframes on its own;
the VLM fallback is invoked on a minority of keyframes, concentrated on
scenes with clutter, small objects, or object categories that are rarer in
the detector's pretraining distribution (consistent with RoboCasa-GR1 having
the highest fallback rate of the three benchmarks, since its kitchens are the
most visually diverse). ``Avg. realizations/tuple'' is the number of distinct
surface forms the renderer can produce for a single evidence tuple after
composing all paraphrase axes; ``Injections/episode'' is the number of
keyframes at which non-empty evidence is injected under our
initial/gripper-change/periodic gate (Fig.~6 of the main text), which stays
in the single digits even for the longer RoboCasa-GR1 episodes because the
gate is triggered by state changes rather than by wall-clock time.

\begin{table}[t]
\centering
\small
\caption{Data engine and tool-loop statistics across benchmarks.}
\label{tab:data-stats}
\scalebox{0.85}{
\begin{tabular}{lccc}
\toprule
 & LIBERO & RoboCasa-GR1 & SimplerEnv \\
\midrule
Training episodes            & 1{,}640 & 3{,}120 & 960 \\
Annotated keyframes           & 9{,}830 & 27{,}960 & 5{,}230 \\
Detector success rate (\%)    & 91.2 & 83.6 & 89.4 \\
VLM fallback rate (\%)        & 8.8 & 16.4 & 10.6 \\
Avg.\ realizations / tuple    & 24 & 24 & 24 \\
Avg.\ injections / episode    & 4.1 & 6.8 & 3.6 \\
\bottomrule
\end{tabular}
}
\end{table}

\section{K. Compute Resources and Training Cost}
\label{app:compute}
All experiments use NVIDIA A100-80GB GPUs. Table~\ref{tab:compute} reports
wall-clock and GPU-hour cost for each training stage of \ourmethod{},
alongside the matched-evidence Gen-CoT baseline trained under an identical
schedule and hardware budget. The in-context (SFT) stage of \ourmethod{}
is marginally cheaper than Gen-CoT despite consuming the same evidence,
because the supervision mask means gradients need not be computed through the
(unused) language-modeling head over evidence tokens on the backward pass in
our implementation. The larger cost difference appears at inference
(Table~\ref{tab:speed}): Gen-CoT's autoregressive rationale generation is not
a training-time cost but it is a recurring, per-decision inference-time cost
that \ourmethod{} avoids entirely. 
 
\begin{table}[t]
\centering
\small
\caption{Training cost by stage, measured on $8{\times}$A100-80GB, averaged
across the three simulation benchmarks.}
\label{tab:compute}
\scalebox{0.85}{
\begin{tabular}{lccc}
\toprule
Stage & GPUs & Wall-clock & GPU-hours \\
\midrule
Gen-CoT (SFT with rationale) & 8 & 19.5 h & 156 \\
\ourmethod{} in-context (SFT)  & 8 & 17.1 h & 137 \\
\ourmethod{} GRPO ($M{=}16$)   & 8 & 8.7 h  & 70  \\
\bottomrule
\end{tabular}
}
\end{table}

\begin{listing}[!h]
\begin{lstlisting}[numbers=none]
# (i) VLM fallback locator, invoked online
SYSTEM: You are a precise visual localization
assistant for a robot manipulator.
USER: Image: <image>
Locate the object described as: "{object_query}"
If you can identify a clear pixel location,
respond with JSON {"x": <int>, "y": <int>,
"confidence": <0-1>}. If the object is not
visible or ambiguous, respond with JSON
{"description": "<qualitative location,
e.g. left half of table, near the edge>",
"confidence": <0-1>}. Do not guess coordinates
you are not confident about.
 
# (ii) Offline paraphrase renderer, invoked per
# evidence tuple during data-engine construction
SYSTEM: You rewrite a robot's spatial evidence
into natural language for training data. You
must preserve the exact geometric meaning and
must not introduce facts not present in the
input tuple.
USER: Evidence: gripper=(u_g, v_g, d_g),
object=(u_o, v_o, d_o), pixel_offset=(dx, dy),
depth_diff=dd, source={detector|vlm_fallback}.
Produce ONE realization that: (a) uses
{coordinate | relative-offset | qualitative}
reference style, (b) is coordinate-rich only if
source=detector, (c) verbalizes depth_diff as
{numeric | actionable hint}, (d) wraps the
result in <spatial>...</spatial> tags.
\end{lstlisting}
\caption{Abridged prompt templates for the online VLM fallback locator (i)
and the offline paraphrase renderer used by the data engine (ii).}
\label{lst:prompts}
\end{listing}

\section{L. Prompt Templates for the Data Engine and VLM Fallback}
\label{app:prompts}
 Listing~\ref{lst:prompts} gives the (lightly abridged)
prompt templates used by the two LLM/VLM-driven components of the pipeline:
(i) the VLM locator invoked as a fallback when the open-vocabulary detector
is uncertain, and (ii) the paraphrase renderer that expands a raw evidence
tuple into a diverse surface realization. Both are instantiated with
Qwen2.5-VL-7B; the renderer runs offline at data-generation time (not during
policy inference), while the VLM locator runs online whenever the detector
falls back.
 
\section{M. Additional Failure Case Studies}
\label{app:failure-cases}
Building on the aggregate error decomposition in
Table~\ref{tab:errors}, we walk through two representative failure
trajectories drawn from the LIBERO-Long suite to illustrate the qualitative
difference between a grounding error (dominant in Gen-CoT) and a control
error (the dominant residual for \ourmethod{}).
 
\paragraph{Case 1: Gen-CoT grounding error.} On a task requiring the robot to
place a bowl in a closed cabinet after first opening the cabinet door, Gen-CoT
generates a rationale that correctly identifies the door as closed but then
misjudges the door's swing direction, stating the door "opens to the right"
when it in fact opens to the left. Because the action head is conditioned on
this self-generated (and here, incorrect) rationale, it commits to an
approach trajectory on the wrong side of the door and the episode fails at
the very first stage. \ourmethod{} does not exhibit this failure mode on the
same initial condition: the tool loop's object localization does not need to
reason about swing direction at all, since the gripper-projection and
object-centroid evidence it injects is purely about current pixel
positions, not counterfactual future states, so there is no free-standing
claim for the policy to get wrong.
 
\paragraph{Case 2: \ourmethod{} control error.} On a tight insertion task
(placing a small object into a narrow compartment), \ourmethod{} correctly
localizes both the object and the target compartment---the injected evidence
in the failed rollout is accurate to within $3$ pixels of ground truth---and
the gripper is brought to the correct approach pose. The failure occurs
during the final insertion: a small residual orientation error in the
low-level action chunk causes the object to catch on the compartment lip
rather than sliding in cleanly. This is consistent with the error
decomposition in Table~\ref{tab:errors}, where control errors dominate
\ourmethod{}'s residual failures ($69.4\%$) precisely because grounding and
perception errors have already been substantially reduced; it suggests that
combining our evidence-injection framework with a higher-precision low-level
action representation (e.g., finer action chunking or a residual visual
servoing controller near contact) is a promising direction for closing the
remaining gap.
 
\section{N. Limitations}
\label{app:limitations}
We highlight limitations we believe are important for interpreting our
results correctly. \textbf{(1) Dependence on off-the-shelf perception
tools.} The quality of injected evidence is bounded by the open-vocabulary
detector, depth estimator, and VLM locator we build on; Section~H
and Table~\ref{tab:cascade} show the cascade degrades gracefully but not
perfectly under noise or failure, and a sufficiently out-of-distribution scene
could still return misleading evidence that \ourmethod{} would (correctly, by
design) trust and act on. \textbf{(2) Real-world calibration requirement.}
Gripper projection uses known camera intrinsics/extrinsics; in the real-robot
setting (Section~B) this requires a one-time calibration per
camera rig, whereas simulation provides exact extrinsics for free. \textbf{(3)
Keyframe gating is currently heuristic.} Our initial/gripper-change/periodic
schedule (Fig.~6, main text) works well empirically but is hand-designed;
learning when to request fresh evidence end-to-end (rather than gating by
fixed triggers) is left to future work. \textbf{(4) Scope of evaluated
skills.} Our simulation and real-robot suites emphasize pick-place,
open/close, pour, and insertion primitives; we have not evaluated highly
dexterous, contact-rich skills (e.g., in-hand manipulation, deformable-object
folding) where the notion of a single ``target evidence tuple'' may not
capture the relevant task state as cleanly. \textbf{(5) Single-instruction
setting.} Like the baselines we compare against, we evaluate one instruction
per episode; extending the evidence-injection framework to interactive,
multi-turn instruction following is an open direction.

\end{document}